\documentclass{article} 
\usepackage[svgnames]{xcolor}
\usepackage{iclr2024_conference,times}

\usepackage{amsmath,amsfonts,bm}

\def\eqref#1{equation~\ref{#1}}

\def\1{\bm{1}}

\DeclareMathAlphabet{\mathsfit}{\encodingdefault}{\sfdefault}{m}{sl}
\SetMathAlphabet{\mathsfit}{bold}{\encodingdefault}{\sfdefault}{bx}{n}

\usepackage{hyperref}
\usepackage{url}

\usepackage{microtype}
\usepackage{graphicx}
\usepackage{subfigure}
\usepackage{subcaption}
\usepackage{booktabs} 
\usepackage{soul}

\usepackage{amsmath}
\usepackage{amssymb}
\usepackage{mathtools}
\usepackage{amsthm}

\usepackage[capitalize,noabbrev]{cleveref}

\theoremstyle{plain}

\theoremstyle{definition}

\theoremstyle{remark}

\usepackage[textsize=tiny]{todonotes}

\newcommand{\ff}{feed-forward}

\newcommand{\nmoe}{N_\text{MoE}}
\newcommand{\nff}{N_\text{ff}}
\newcommand{\dff}{d_\text{ff}}
\newcommand{\nact}{N_\text{act}}
\newcommand{\dexp}{d_\text{expert}}

\newcommand{\dexpert}{d_\text{expert}}
\newcommand{\dmodel}{d_\text{model}}
\newcommand{\nexpert}{N_\text{expert}}
\newcommand{\ntokens}{n_\text{tokens}}
\newcommand{\nblocks}{n_\text{blocks}}
\newcommand{\nlayers}{n_\text{layers}}

\newcommand{\nheads}{n_\text{heads}}

\definecolor{purp}{RGB}{21,171,0}

\title{Scaling Laws for Fine-Grained\\ Mixture of Experts
}

\author{Jakub Krajewski $^{\ast}$ \\ University of Warsaw \\ IDEAS NCBR \And Jan Ludziejewski $^{\ast}$ \\ University of Warsaw \\ IDEAS NCBR
 \And Kamil Adamczewski \\ IDEAS NCBR  \And Maciej Pióro\hspace{1.1cm} \\ IPPT PAN \\ IDEAS NCBR \AND Michał Krutul \\ University of Warsaw \\ IDEAS NCBR \And Szymon Antoniak \\ University of Warsaw \\ IDEAS NCBR \And Kamil Ciebiera \\ University of Warsaw \\ IDEAS NCBR \And Krystian Król \\ University of Warsaw \\ IDEAS NCBR \AND Tomasz Odrzygóźdź \\ TradeLink \And Piotr Sankowski \\ University of Warsaw \\ IDEAS NCBR \And Marek Cygan \\ University of Warsaw \\ Nomagic \And Sebastian Jaszczur $^{\ast}$ \\ University of Warsaw \\ IDEAS NCBR
}

\iclrfinalcopy 
\begin{document}

\maketitle

\begin{abstract}

Mixture of Experts (MoE) models have emerged as a primary solution for reducing the computational cost of Large Language Models. In this work, we analyze their scaling properties, incorporating an expanded range of variables.  Specifically, we introduce a new hyperparameter, granularity, whose adjustment enables precise control over the size of the experts. Building on this, 
we establish scaling laws for fine-grained MoE, taking into account the number of training tokens, model size, and granularity.  Leveraging these laws, we derive the optimal training configuration for a given computational budget. Our findings not only show that MoE models consistently outperform dense Transformers but also highlight that the efficiency gap between dense and MoE models widens as we scale up the model size and training budget. Furthermore, we demonstrate that the common practice of setting the size of experts in MoE to mirror the \ff{} layer is not optimal at almost any computational budget. 
\noindent\let\thefootnote\relax\footnote{Contributions: Jakub implemented fine-grained MoE, ran experiments, and oversaw the course of the project. Jan designed and implemented the scaling laws, also optimized and tuned the fine-grained MoE implementation.
Kamil A. provided significant advice on many aspects of the project. Maciej experimented with the block design and, with Michał, provided considerable technical support. Szymon, Kamil C., Krystian, and Tomasz contributed to the project and the engineering in various ways. Marek, along with Piotr, provided high-level scientific advice. Sebastian came up with the initial idea, started the project, and supervised it while setting the research direction and leading experiments and analyses. Correspondence to \textless{}s.jaszczur@uw.edu.pl\textgreater{}. $^{\ast}$~Equal contribution. 
}

\end{abstract}

\section{Introduction}
In recent years, we have witnessed Large Language Models (LLMs) achieve exceptional performance in tasks across numerous domains \citep{chowdhery2022palm, yin2023survey, agostinelli2023musiclm}. However, training those massive models incurs high computational costs, measured in millions of GPU-hours \citep{touvron2023llama2}, enabled only by enormous budgets \citep{workshop2023bloom} and leading to non-negligible carbon footprints \citep{faiz2024llmcarbon}. To combat these obstacles, the research community has been striving to increase the efficiency of LLMs. One promising approach that has lately been gaining visibility is the use of Mixture of Experts (MoE) methods.
Models such as Switch \citep{fedus2022switch} and Mixtral \citep{jiang2024mixtral} have already demonstrated that it is possible to achieve comparable effectiveness with significantly lower computational costs.


In the context of the current trend of increasing budgets for training language models, a question arises: will MoE models continue to be attractive in the future? This is an important issue, as other studies have stated that the gap in efficiency between MoE and standard Transformers narrows at scale \citep{artetxe2022efficient} or even that traditional dense models may outperform MoE as the size of the models increases \citep{clark2022unified}.

In this paper, we argue that previous claims lose their validity when we relax certain implicit assumptions regarding the training process, present in previous research. In particular, we refer to the fixed training duration and the constant size of experts in MoE models. 

\begin{figure*}[t]
  \centering
  \includegraphics[width=0.99\textwidth]{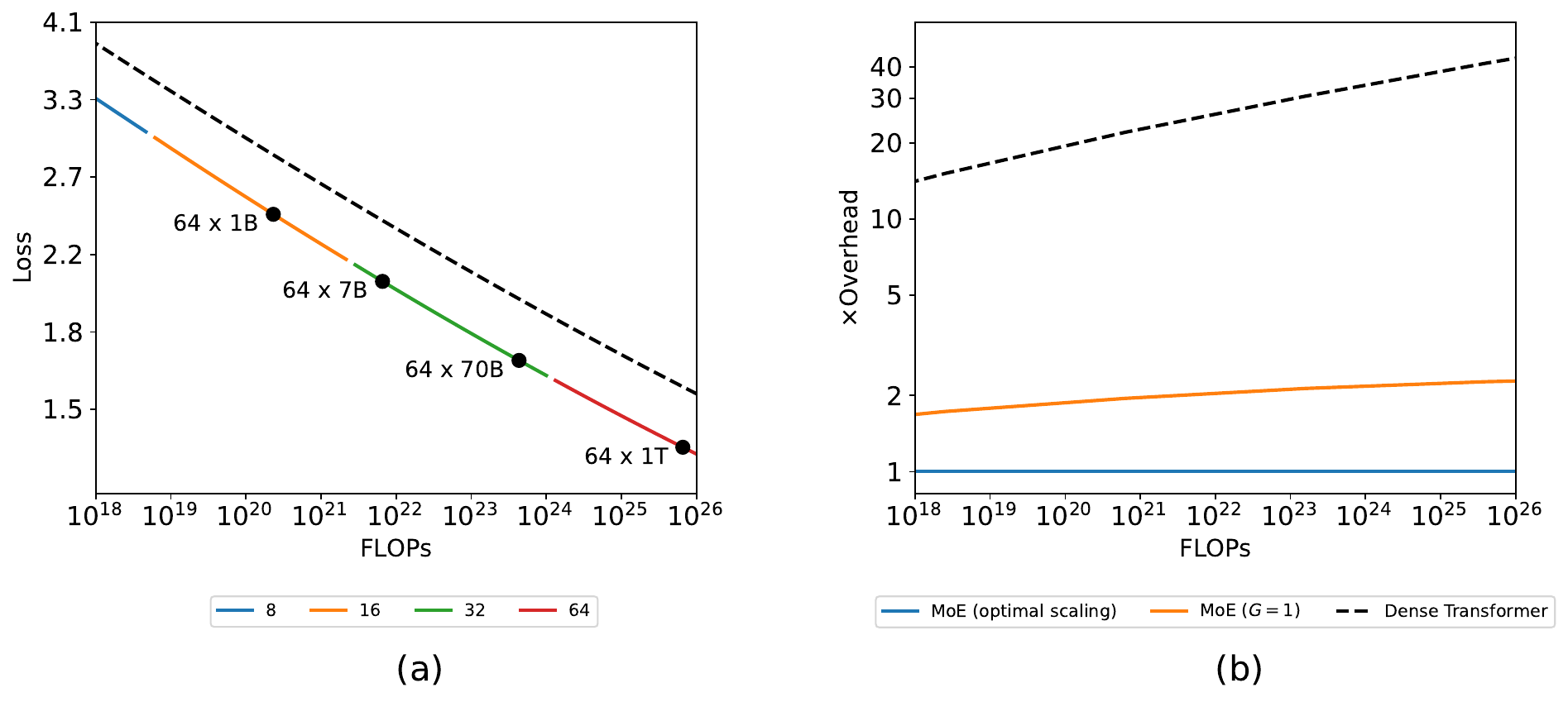}
  \caption{Mixture-of-Experts can be \textit{always} considered more efficient than dense Transformers, regardless of the model size. \textbf{(a)} Compute Optimal scaling curves for MoE and standard Transformers. The dashed line represents a dense Transformer. Colors denote optimal granularity for the given FLOPs training budget. \textbf{(b)} Relative number of FLOPs needed to train Transformer and Vanilla MoE (MoE with $G=1$) to achieve the performance of MoE with compute optimal $G$.}
  \label{flag_plots}
\end{figure*}

Our results suggest that a compute-optimal MoE model trained with a budget of $10^{20}$ FLOPs will achieve the same quality as a dense Transformer trained with a \textit{$20\times$} greater computing budget, with the compute savings rising steadily, exceeding \textit{$40\times$} when budget of $10^{25}$ FLOPs is surpassed (see Figure \ref{flag_plots}). Importantly, we show that the standard practice of fixing the size of experts in MoE to be the same as feed-forward layer is \emph{almost never} optimal.

Our main contributions are:

\begin{enumerate}
    \item Introducing a new hyperparameter - granularity. Adjusting this parameter allows us to determine the optimal size of experts in MoE models, which translates into increased efficiency.

    \item Deriving new scaling laws for MoE models that incorporate variable training duration, the number of parameters, and granularity. Such scaling laws allow us to calculate optimal training hyperparameters for MoE models.
    


\item Demonstrating that, with optimal settings, MoE models can always outperform traditional Transformers at any computing budget. This is a conclusion contrary to the results from \cite{clark2022unified}.
\end{enumerate}

The code used to produce the results described in this work is open-sourced at \href{https://github.com/llm-random/llm-random}{\texttt{github.com/llm-random/llm-random}}.

\section{Related Work }
\textbf{Mixture of Experts. \ \ } 
In the context of language modeling, MoE was first introduced by \cite{shazeer2017outrageously} as a sparsely gated layer between stacked blocks of LSTM \citep{hochreiter1997long}. A similar technique was proposed in the context of Transformers by \cite{shazeer2018meshtensorflow} and \cite{lepikhin2020gshard}. \cite{fedus2022switch} proposed to route each input to only a single expert and designed a modified initialization scheme to reduce training instability.
Numerous studies have proposed to modify the original routing method. \cite{lewis2021base} used a linear assignment algorithm to postprocess token-expert mappings and ensure even expert selections. \cite{roller2021hash} suggested another approach involving deterministic hash functions. \cite{zhou2022mixtureofexperts} proposed expert choice routing, eliminating the need for additional load balancing losses. \cite{puigcerver2023sparse} designed a fully-differentiable Soft MoE architecture.

Concurrently to our work, \cite{dai2024deepseekmoe} proposed to modify the MoE layer by segmenting experts into smaller ones and adding shared experts to the architecture. Independently, \cite{liu2023unified} suggested a unified view of sparse \ff{} layers, considering, in particular, varying the size of memory blocks. Both approaches can be interpreted as modifying granularity.  However, we offer a comprehensive comparison of the relationship between training hyperparameters and derive principled selection criteria, which they lack.

\textbf{Scaling laws.} \ Scaling laws are empirically derived equations relating the loss of a model with variables such as the number of parameters, training samples, or the computational budget. In the case of dense Transformers, scaling laws were first studied by \cite{kaplan2020scaling}, who observed power law relationships between the final model perplexity and model and dataset size. This work was extended by \cite{hoffmann2022training} by considering variable cosine cycle lengths and formulating a modified functional form of the scaling equation.

Scaling laws have also been proposed for other architectures and training scenarios. \cite{henighan2020scaling} studied autoregressive modeling across various modalities, while \cite{ghorbani2021scaling} considered machine translation. \cite{frantar2023scaling} explored the impact of pruning on vision and language Transformers, deriving optimal sparsity for a given compute budget. \cite{clark2022unified} studied the scaling of MoE when changing model size and number of experts on a fixed dataset, concluding that routed models are more efficient only until a certain model size. In this work, we challenge that claim by considering a variable, optimal dataset size for both model families (see Section~\ref{moe_more_eff}).


\section{Background}
\label{background}

\subsection{Model Architecture}

\paragraph{Transformer.} A standard decoder-only Transformer \citep{radford2018improving, noauthororeditor, kaplan2020scaling, brown2020language} consists of an embedding layer, a stack of alternating attention and \ff{} layers, and an unembedding layer. In the model, each input token is converted by the embedding layer into a vector of size $\dmodel$, the dimension maintained across all the layers in the residual stream.

The \ff{} component consists of two linear transformations and a nonlinearity $\phi$ in between. It can be described as
    $\text{FFN}(x) = \phi(xW_1+b_1)W_2+b_2$,
with $W_1$ mapping from $\dmodel$ to $\dff$, and $W_2$ back to the original $\dmodel$. It is standard \citep{radford2018improving, rae2022scaling, touvron2023llama, jiang2023mistral} to set the hidden dimension as $\dff = 4 \cdot \dmodel$. 

Feed-forward layers contain the majority of Transformer parameters and require the biggest computational budget counted in terms of FLOPs. Subsequently, they are the main focus of the Mixture of Experts models considered in this work.

\paragraph{Mixture of Experts.} 
The core idea behind MoE in Transformers is to replace the feed-forward layer with a set of $\nexpert$ \textit{experts}. The size of each expert is typically \citep{fedus2022switch, zhou2022mixtureofexperts, zhou2023brainformers, jiang2024mixtral} set to mirror the original dimensions of the layer, with the hidden expert dimension $\dexp{}$ equal to $\dff.$ Therefore, the total number of parameters in MoE scales linearly with the number of experts. 
However, the computational cost remains approximately constant as each input is routed and then processed by a subset of experts.

\newpage
\subsection{Scaling Laws}
\label{sec:back_scaling}

\textbf{Dense Transformers.} Large Transformer-based models are known to approximately obey the power-law relationship between final loss $\mathcal{L}$, model size $N,$ and number of training tokens $D.$ This relationship is often called {\it Chinchilla scaling laws} described by \cite{hoffmann2022training} as
\begin{align}\label{eq:chinchilla}
    \mathcal{L}(N, D) = c + \frac{a}{N^\alpha} + \frac{b}{D^\beta}. 
\end{align}

The power-law formula is composed of three distinct terms that characterize the intrinsic entropy of data, constraints of the model, and limitations in the training data. The term $c$ represents the minimum possible error intrinsic to the data. The remaining two terms are suboptimality terms, which address the limitations in function representation owing to the size of the model and in data signified by the number of tokens. In the limit, with infinite data and model size, the loss is reduced to $c$.


\textbf{Mixture of Experts.}  For MoE Transformer-based models, \cite{clark2022unified} formulated the final loss for a constant dataset size \( D \) of 130B tokens, allowing for variations in the expansion rate \( E \), as: 
\begin{align}\label{eq:scaling_law}
    \mathcal{L}(N, E) = \left( \frac{10^{d/a}}{N}\right)^a \left( \frac{1}{E}\right)^{b+c\log{N}}.
\end{align}

However, this result has a notable limitation as it can be applied only to the original dataset size. The scalability and effectiveness are constrained in this scenario because it is crucial to align the number of training samples with the available computational resources for optimal use. As per \cite{kaplan2020scaling} and \cite{hoffmann2022training}, maintaining a constant dataset size while scaling up the neural network size leads to undertraining, resulting in a model that does not perform to its full potential.

\section{Granularity}

As described in Section~\ref{background}, in the standard setting, the inner dimension of each expert network, $\dexp$, is equal to $\dff$, which is the same size as the \ff{} layer of the base model.
\begin{figure*}[t]
  \centering
  \begin{tabular}{ c @{\hspace{30pt}} c }
    \includegraphics[width=.45\textwidth]{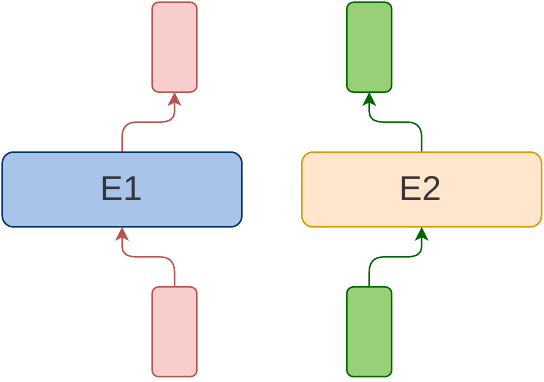} &
      \includegraphics[width=.45\textwidth]{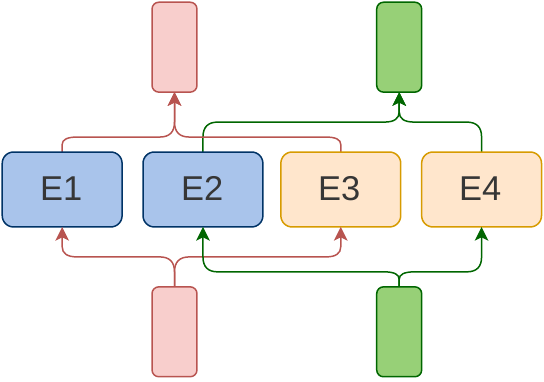} \\
    \small \textbf{(a)} &
      \small \textbf{(b)}
  \end{tabular}

  \medskip

  \caption{\textbf{(a)} Standard MoE layer with $G=1$ \textbf{(b)} Corresponding MoE layer with $G=2$. Each of the original experts is split into two granular ones. The split occurs in the hidden dimension of an expert. Increasing $G$ allows for a more precise mapping between experts and tokens. Since for granularity $G$, the token is routed to $G$ granular experts, the number of parameters activated per token is the same in both cases.}
  \label{diag_gran_better}
\end{figure*}

In this work, we suggest an alternative approach where the hidden dimension of the expert is not necessarily set to mirror that of the standard feed-forward layer. Instead, it can be adjusted to a value that is the most effective. This approach allows the configuration of MoE to be articulated in terms of two key hyperparameters: \textit{granularity} (\(G\)) and \textit{expansion rate} (\(E\)). In the following parts of this work, we will also use the term \textit{active} parameters to refer to the non-embedding parameters used to produce output for a single token, except routing. The number of active parameters is denoted as $\nact$.

Let $\dexpert$ be the hidden dimension of a single expert. Granularity is defined as
\begin{align*}
    G = \frac{\dff}{\dexpert}.
\end{align*}
In other words, granularity denotes the multiplier factor for the change in the size of an expert from the original standard model, defined as $G=1$. 
In this work, we investigate $G > 1$ where experts are smaller than in the standard layer. 

Note that increasing granularity does not affect the number of active parameters. As $G$ increases, the number of experts that process the token grows proportionally to $G$. In other words, for granularity $G$, a token is routed to $G$ fine-grained experts, thereby keeping the number of active parameters constant. See Fig. \ref{diag_gran_better} for visualization.

We then define the \textit{expansion rate}, which describes the increase in the number of parameters from a standard transformer layer to a MoE layer. Given that, $\nmoe$ and $\nff$ denote the total number of parameters in a MoE layer excluding routing and the standard \ff{} layer, respectively. The expansion rate $E$ is then defined as 
\begin{align*}
    E = \frac{\nmoe}{\nff}.
\end{align*}

Expansion rate can also be seen as the total number of parameters in a MoE layer compared to its active parameters. 

The concept of the expansion rate is intricately linked to the number of experts through the idea of granularity. Indeed, the definitions of both granularity and expansion rate extend and refine our understanding of the number of experts, symbolized as $\nexpert$.
\begin{align}
    \nexpert = G \cdot E
\end{align}

For non-granular models, where $G=1$, the expansion rate is equal to the number of experts.

Intuitively, increasing granularity for a given expansion rate gives the model more flexibility in mapping datapoints to experts, potentially improving performance. We incorporate the notion of granularity into our scaling laws in Section~\ref{sec:scaling_laws}. The discussion about practical tradeoffs in changing this parameter is given in Section~\ref{sec:scaling_app}.

\section{Scaling Laws}\label{sec:scaling_laws}


Granularity determines changes in the architecture of MoE. In this section, we answer a central question of this work: whether the granular MoE models follow scaling laws and, if so, how granularity affects them. Thus, we aim to derive a parametric scaling law for predicting the final loss value $\mathcal{L}$ based on granularity $G$, total number of non-embedding parameters $N$, and number of training tokens $D$.

We run over 100 experiments on the decoder-only Transformer architecture, with each \ff{} component replaced by a Mixture of Experts layer. Those experiments involve training models with sizes ranging from 129M to 3.7B parameters across different training durations, from 16B to 130B tokens. We consider logarithmically spaced values of granularity between 1 and 16. To constrain the search space, $E=64$ is fixed, following the recommendations of \cite{clark2022unified}.  In addition, we also run experiments with dense Transformers to compare their performance with MoE. The details of all architectures, the training procedure, and hyperparameter choices are described in detail in Appendix~\ref{app:setup}.

In the subsequent part of this paper, we will use the notation $E\times \nact$ to describe a MoE model with $\nact$ active parameters and expansion rate $E.$

\newpage
\subsection{Power Law With Respect to Granularity} 
\begin{figure*}[t]
  \centering
  \includegraphics[width=.99\textwidth]{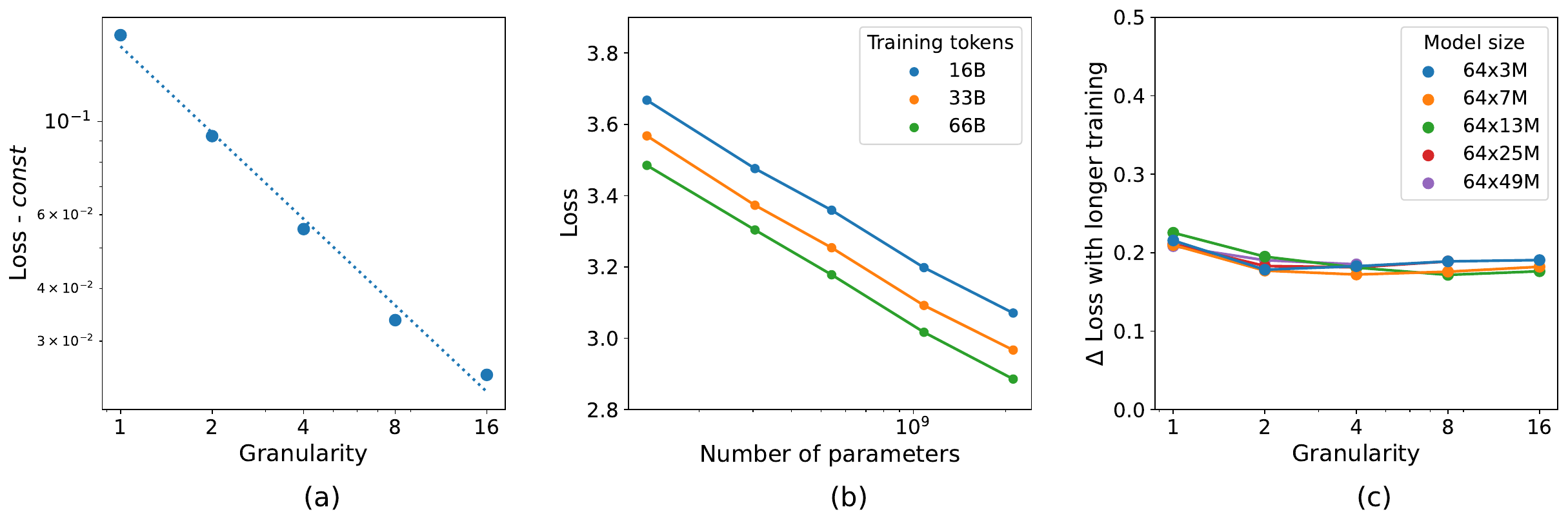}
  \caption{\textbf{(a)} The effect of $G$ on $\mathcal{L}_{N,D}(G)$ for constant $N$ and $D$. Both axes are in the log-scale. The results suggest the linear relationship between $\log(G)$ and $\log(\mathcal{L}-c)$.
The given values are $N=64 \times 25M$, $D=16B$, $const=3.12$ .  The plots for additional values of $N$ and $D$ can be found in Appendix \ref{sec:app_add_viz}. \textbf{(b)} The impact of varying the number of parameters $N$ on the loss for fixed granularity $G=4$. For other granularity values, see Appendix \ref{sec:app_add_viz}. \textbf{(c)} The difference in the loss between training for 16B and 65B tokens for all model sizes and granularity values. The model size is reported as the expansion rate and the number of active parameters.}
  \label{empirical}
\end{figure*}

We first answer the question of whether granular models follow the scaling laws. In Figure~\ref{fig:scalinglaws}(a), it can be seen that increasing granularity results in a lower loss. The returns follow approximately an exponential pattern, converging to a positive constant.   The empirical relationship given by Figure \ref{empirical}(a) suggests the following power-law dependence of loss on a varying granularity for given $N$ and $D$ and constants $g, h$ and $\gamma$ that may be dependent on them,
\begin{align}\label{eq:g_only}
    \mathcal{L}_{N,D}(G) = \frac{g_{N,D}}{G^{\gamma_{N,D}}} + h_{N,D}.
\end{align}
\subsection{Scaling the Model and Dataset Size}
\label{sec:scaling_model_data}
As outlined in Section~\ref{sec:back_scaling}, the power-law given by Eq.~\ref{eq:chinchilla} consists of three terms that describe inherent data entropy and limitations in function representation and data. This derivation is independent of the architecture. In particular, the Eq.~\ref{eq:chinchilla} also holds for constant granularity. Empirically, we observe a power law relationship in $N$ and $D$ analogous to that in dense models as depicted in Figure~\ref{empirical}(b) for a fixed value of granularity (see also Fig.~1, \cite{kaplan2020scaling}).  Furthermore, the validity of this functional form is verified by fit in Section~\ref{sub:parametric}. 

Since we know that separate scaling laws are valid for given granularities, in the general form, the parameters in Eq.~\ref{eq:chinchilla} can be dependent on the model's granularity:
\begin{align}\label{eq:gran_func}
     \mathcal{L}_G(N, D) = c_G + \frac{a_G}{N^{\alpha_G}} + \frac{b_G}{D^{\beta_G}}.
\end{align}



\subsection{The Form of the Joint Scaling Law}\label{sub:joint}

Following the above observation that models with constant granularity obey Chinchilla scaling laws given by Eq.~\ref{eq:chinchilla}, the key question arises as to how the general notion of granularity $G$ can be incorporated into the joint scaling law. 
Moreover, the scaling law formula from Eq.~\ref{eq:gran_func} for constant $N$ and $D$ has to be representable by Eq. \ref{eq:g_only}. This is because the former is a more general equation, encompassing shared hyper-parameters across all $N$, $D$, and $G$. It is anticipated to align with the latter, consisting of distinct power laws, each with specific parameters for different $N$ and $D$ values. Consequently, the objective is to identify a function that fulfills these criteria.

\begin{align}\label{eq:l_general}
     \mathcal{L}(N, D, G) &= \hspace{0.43cm} \mathcal{L}_{N,D}(G)  \hspace{0.43cm} = \hspace{0.6cm} \mathcal{L}_G(N,D) & \\
      &= \frac{g_{N,D}}{G^{\gamma_{N,D}}} + h_{N,D} = c_G + \frac{a_G}{N^{\alpha_G}} + \frac{b_G}{D^{\beta_G}} & \nonumber
\end{align}

In the subsequent sections, we aim to determine which of these parameters remain independent of $G$ and identify their functional form. Furthermore, we present some rationale for the structure of our formula.

\textbf{Lower Bound.}
Consider the limit of Eq.~\ref{eq:gran_func} for $N$ and $D$ growing to infinity:
\begin{align}\label{eq:gran_limit}
\lim_{\substack{N\to \infty\\ D\to \infty}}    \mathcal{L}(N, D, G) = c_G.
\end{align}
 with the constant term $c_G$ dependent on granularity. This is contradictory to the fact that it captures the inherent entropy of the dataset. Lower bound of the achievable loss when training bigger models on more samples should not depend on the architecture, therefore parameter $c_G=c$ is constant for all granularities. 

\textbf{Granularity and Number of Tokens $D$.} 
As seen in Figure~\ref{empirical}(c), the benefit of training a model on a larger dataset is almost the same for each granularity value. This suggests that there is no interaction between $D$ and $G$. Therefore, we can assume that 

 \begin{align}
     \frac{b_G}{D^{\beta_G}} = \frac{b}{D^\beta}.
\end{align}

\textbf{Granularity and Model Size $N$.}  
We consider \( \alpha \) to be a constant that describes how the function scales with \( N \). In this work, we assume polynomial functional forms that rule out the potential dependency of $\alpha$ on $G$ given the form of Eq. \ref{eq:g_only}. 
Therefore, 
the only element dependent on $G$ is $a_G$:
\begin{align}\label{eq:loss_moe}
    \mathcal{L}(N, D, G) = c + \left(\frac{g}{G^\gamma} + a\right)\frac{1}{N^\alpha} + \frac{b}{D^\beta}.
\end{align}
Finally, one could consider omitting the constant $a$ in the equation above, and it would still reduce to \ref{eq:g_only} for constant $N$ and $D$. However, this would mean that a model with infinite granularity and a small number of active parameters can achieve the perfect perplexity of the lower bound.  We assume that a sparse MoE (Mixture of Experts) model is unlikely to surpass the performance of an equivalent dense model that has a matching total number of parameters, all of which are active. This means that constant $a$ can act as a marginal improvement due to granularity.

\begin{figure*}[t]
  \centering
  \includegraphics[width=.99\textwidth]{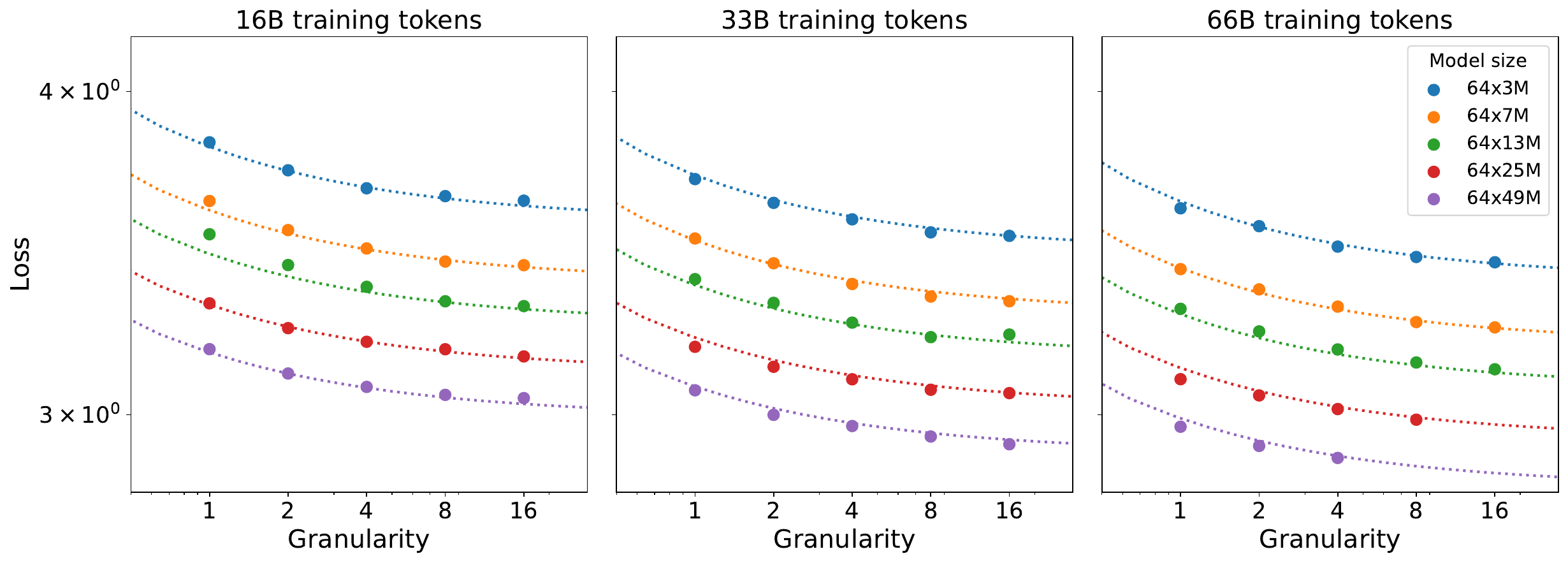}
  \caption{Fit of the scaling laws compared to the experimental results.}
  \label{fig:scalinglaws}
\end{figure*}
Subsequently, we fit parameters in Eq. \ref{eq:loss_moe} to describe the scaling of MoE. For comparison, we also perform fitting for dense transformer given by Eq.~\ref{eq:chinchilla}. Similarly to \cite{hoffmann2022training}, we use Huber loss \citep{huber_loss}, with $\delta=0.1$. The optimization is performed using the BFGS algorithm. We include a weight decay of $5e-4$ to enhance generalization. We start with fitting parameters in Eq.~\ref{eq:loss_moe} and then find architecture-dependent coefficients $\alpha, \beta, A$ and $B$ in Eq.~\ref{eq:chinchilla}. We observe a good fit, with $\text{RMSE}=0.015$. The values are presented in Table \ref{tab:coeffs_scaling}. We depict the results in Figure~\ref{fig:scalinglaws}. 

\newpage
\subsection{Fitting the Parametric Scaling Law}\label{sub:parametric}
\begingroup
\begin{table}[h]
    \centering
        \setlength{\tabcolsep}{5pt} 
    \caption{Values of the fitted coefficients.}
    \vspace{0.2cm}
    \begin{tabular}{ccccccccc}
        \toprule
        Model  & a & $\alpha$ & b & $\beta$ & g & $\gamma$ & c \\
        \midrule        
        MoE  & 18.1 & 0.115 & 30.8 & 0.147 & 2.1 & 0.58 & 0.47 \\ 
        Dense & 16.3 & 0.126 & 26.7 & 0.127 & - & - & 0.47\\
        \bottomrule
    \end{tabular}
    \label{tab:coeffs_scaling}
\end{table}
\endgroup

\begin{figure*}[t]
  \centering
  \includegraphics[width=.9\textwidth]{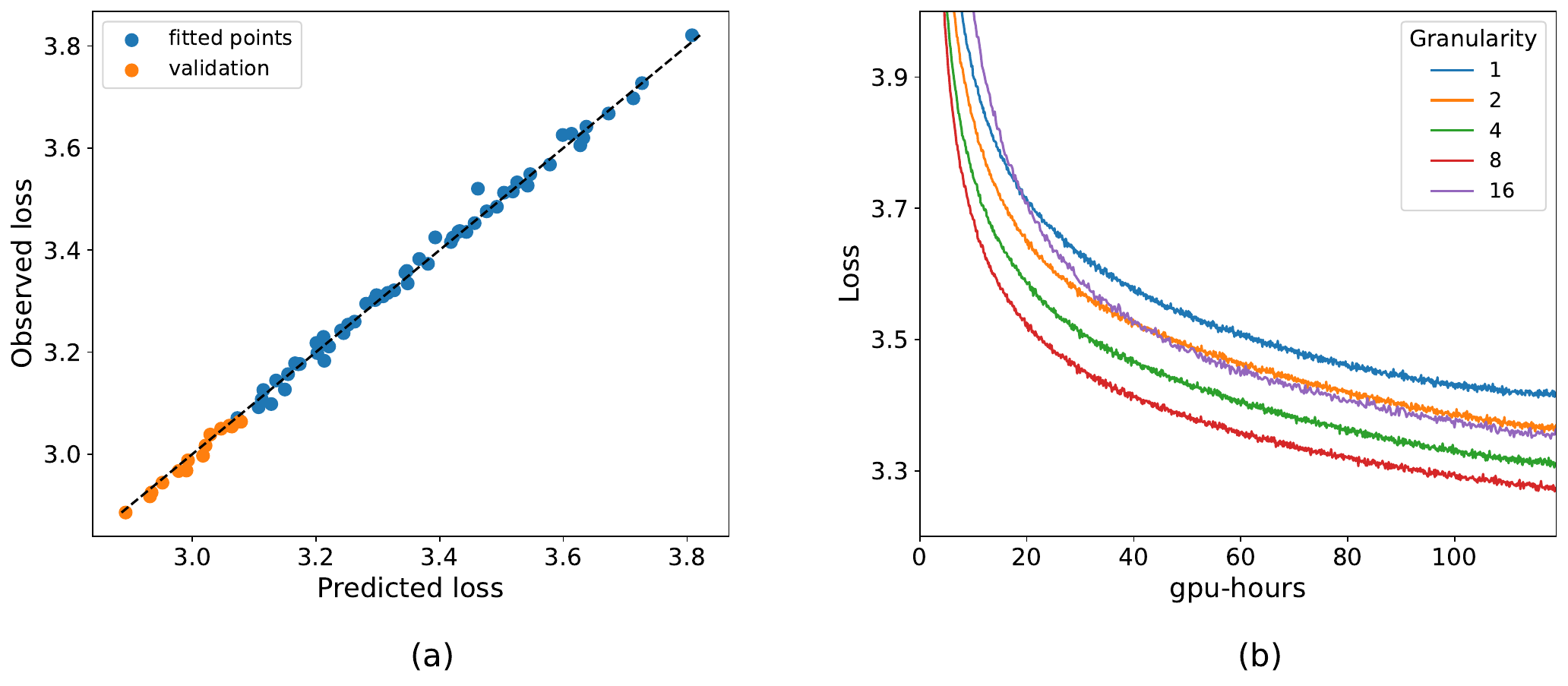}
  \caption{\textbf{(a)} Validation of the scaling laws. \textbf{(b)} Training loss curves for model with $N=64 \times 7M$, $D=66B$ tokens, measured against wall-clock time on NVIDIA A100 GPU. $G=8$ leads to the best performance, as for $G=16$ the routing cost dominates gains from granularity. We model the increased cost of routing by measuring FLOPs for each configuration.}
  \label{fig:valid_wall}
\end{figure*}

We validate the stability of the fit by excluding the top $20\%$ of models with the lowest perplexity and finding the coefficients based on the remaining experiments. We observe that the formula remains almost unchanged in this scenario (see Table \ref{tab:coeffs_valid} in Appendix \ref{app:coeffs_valid}). The validation RMSE is 0.019. Results are depicted in Figure~\ref{fig:valid_wall} (a).  

\subsection{MoE Scaling Properties}\label{sub:scaling_properties}
Comparing the part of the formula that approximates underfitting (that is, dependent on training tokens) in MoE ($30.8D^{-0.147}$) and Transformer ($26.7D^{-0.127}$), we can infer that MoE models need longer training to perform competitively but scale better after reaching that point. Nonetheless, this moment may still precede the compute optimal for both models.
On the other hand, we can see that the exponent on dense models $\alpha=-0.126$ scales better with a total number of parameters than the MoE counterpart $\alpha=-0.115$. This should not be surprising since dense models use all parameters on each token contrary to MoE, which gains a computational advantage by activating only a subset of them. Therefore, the fair comparison of the performance has to take into account FLOPs used by each model type. In the next section, we find compute-optimal granularity for a given FLOP budget.

\section{Optimal Allocation of Computational Budget }\label{sec:scaling_app}

In Section \ref{sec:scaling_laws}, we show that higher granularity leads to lower loss for the same number of training steps. This is not always the case if we consider the wall-clock time. As depicted in Figure \ref{fig:valid_wall} (b), in practice for too high values of $G$ (relative to $\dmodel$), training can be bottlenecked by the routing cost. Practical modeling of this situation is possible by measuring FLOPs in routing.
In this section we find optimal $N,D,G$ for a given computational budget $F$ 
by solving the following optimization problem,
\begin{equation*}
    \begin{aligned}
        & \underset{N, D, G}{\text{minimize}}
        & & \mathcal{L}(N, D, G) \\
        & \text{subject to}
        & & \text{FLOPs}(N, D, G) = F.
    \end{aligned}
\end{equation*}

\subsection{Computational Cost of Granularity} \label{sub:compute_optimal}


It is important to acknowledge that increasing granularity can lead to some challenges in training the model, namely higher computational and communication costs and a larger memory footprint. The main component responsible for higher costs is the increase in routing operations due to a larger pool of granular experts. This increase is proportional to the value of $G.$ For standard, non-granular MoE models ($G=1$), the routing overhead still exists, although it has been considered negligible. 

Taking into account the routing operation overhead, the number of used FLOPs $F$ is described by the following formula:
\begin{align}
    F = (12{\dmodel}^2c_f + \dmodel EGc_r) \cdot D \cdot \nblocks,
    \label{eq:flops}
\end{align}
given expansion rate $E$, granularity $G$, and constants that denote FLOPs per active parameter ratio, respectively, within routing ($c_r$) and within the rest of the network ($c_f$). The term $12{d_{\text{model}}}^2$ is the number of active parameters within a transformer block, while $d_{\text{model}}EGc_r$ is the number of active parameters within a routing network. The in-depth analysis of constants $c_r$ and $c_f$ can be found in Appendix
\ref{app:constants}. We exclude embedding and unembedding from the FLOPs calculations, following \cite{hoffmann2022training}.

Observe that, in contrast to scenarios where routing operations are omitted, the FLOPs calculation that incorporates routing overhead relies on both $d_{\text{model}}$ and $\nblocks$. Consequently, an additional condition is required to determine the scaling of $d_{\text{model}}$ and $\nblocks$ in relation to an increase in $N$, the number of parameters. It is noted that minor variations in the depth-to-width ratio are not significant \citep{kaplan2020scaling}. Following this analysis, we opt to adopt the assumption that $\dmodel=64\nblocks$.

The total number of parameters in the feed-forward layer, excluding the routing matrix, is $2E\dff d_{\text{model}} = 8E{\dmodel}^2$, and $4{\dmodel}^2$ in attention (key, query, value, and output projection). This results in the following formula for the total number of parameters, $N = {\dmodel}^2 \cdot (8E + 4) \cdot \nblocks$.






\subsection{Compute Optimal Formula} \label{optimi}

Taking into consideration we need to solve the following optimization problem, given $F$,

\begin{equation*}
    \begin{aligned}
        & \underset{N, D, G}{\text{minimize}}
        & & \mathcal{L}(N, D, G) \\
        & \text{subject to}
        & & F = (12{\dmodel}^2c_f + \dmodel EGc_r) \cdot D \cdot \nblocks\\
        & & & N = d_{\text{model}}^2 \cdot (8E + 4) \cdot n_{\text{layers}}, \\
        & & & d_{\text{model}} = 64 \cdot n_{\text{layers}}.
    \end{aligned}
\end{equation*}

All these constraints are reducible to a one-dimensional optimization problem, which is, however, hard to solve analytically. Therefore we approximate the solution using Brent's method \citep{Brent1971AnAW}. The results of this optimization for varying FLOPs budgets are plotted in Figure \ref{flag_plots} while the optimal configurations of parameters for selected model sizes are presented in Table \ref{compute_opt}. To validate the uncertainty of these predictions, we follow \cite{hoffmann2022training} and calculate the 10$\text{th}$ and 90$\text{th}$ percentiles estimated via bootstrapping data (see Appendix \ref{app:reliabiliy} for the detailed results).

\begin{table}[h]
    \centering
    \caption{Compute optimal training hyper-parameters for MoE models. Optimal $N$ and $D$ follow approximately similar relation to these of \cite{hoffmann2022training} for active parameters around the range of $1B$ to $10B$ parameters, requiring comparably longer training for smaller models and shorter for bigger ones. Higher granularity is optimal for larger compute budgets.
    }
    \vspace{0.2cm}
    \begin{tabular}{ccccc}
    \toprule
        N & D & G & FLOPs & Loss \\
        \midrule
        64 x 100M & 4.37B  & 8 & 2.95e+18 & 3.133 \\ 
        64 x 1B & 28.94B  & 16 & 1.93e+20 & 2.491 \\
        64 x 3B & 72.90B  & 16 & 1.41e+21 & 2.245\\
        64 x 7B & 137.60B  & 32 & 6.46e+21 & 2.076 \\
        64 x 70B & 941.07B  & 32 & 4.16e+23 & 1.694  \\
        64 x 300B & 2.96T & 64 & 5.69e+24 & 1.503  \\
        64 x 1T & 7.94T  & 64 & 4.97e+25 & 1.367 \\
        \bottomrule
    \end{tabular}
    \label{compute_opt}
\end{table}

\subsection{MoE is Always More Efficient} \label{moe_more_eff}
Contrary to the results from \cite{clark2022unified}, in Figure \ref{flag_plots} we can see, that Mixture-of-Experts can be always considered more efficient than dense Transformers, regardless of the model size. According to our previous observations from Section~\ref{sub:scaling_properties}, MoE models scale better with optimal training. However, for short training schedules, they may under-perform dense models. This means that for constant training time and increasing model size, there exists a point where both models will become very under-trained, in which scenario dense models surpass MoE. This shows why in \cite{clark2022unified}, where varying the number of training tokens has not been considered, MoE was predicted to be under-performing for models bigger than $1T$. However, when all training hyper-parameters $N,D,G$ are properly selected to be compute-optimal for each model, the gap between dense and sparse models only increases as we scale.



\section{Discussion}\label{sec:analysis}

\paragraph{Extreme Granularity.}
In Section \ref{sec:scaling_laws}, we argue that model performance improves with increasing granularity. This postulate largely aligns with the empirical findings of our study. Nonetheless, at exceedingly high granularity levels, such as $G=64$ in models characterized by $\dmodel=256$  and $E=64$, there is an observable decline in performance. This phenomenon is particularly evident in scenarios where the number of parameters in the routing mechanism exceeds active parameters in actual experts. Additionally, as described in Section \ref{sec:scaling_app}, the utility of such high granularity is predominantly restricted to models of substantial size. In alignment with the principles outlined by \cite{hoffmann2022training}, this research focuses more on findings that can be broadly applied rather than delving into the specific details of these corner-case situations. However, it is hypothesized that the efficiency of models with significantly high granularity could be potentially enhanced through careful expert initialization or modifications to the routing algorithm. These ideas are set aside to be investigated in future studies.

\paragraph{Varying Expansion Rate.}
In this study, due to computational resources constraint, we focus on $E=64,$ as recommended by \cite{clark2022unified}. This value of $E$ was also used for the largest models in other works \citep{du2022glam, zhou2022mixtureofexperts} and the best-performing configuration in \cite{fedus2022switch}. Nonetheless, we acknowledge the importance of considering different expansion rates, as different levels of $E$ may be chosen based on factors like the target size of the model in memory. Therefore, in Appendix~\ref{app:exp_rate}, we present the results of the study for $E=16$ and show that the main findings of this work are still valid in such cases. 

\paragraph{Including $E$ in the formula.} Another possible advancement would be to unify all of the factors $N, D, G$ and $E$ in one formula. While this would open the possibility of studying the relationships between coefficients in more detail, it would also be hard to practically recommend the optimal configuration in such a scenario using only FLOPs. This is because larger values of $E$ typically lead to better performance but also incur additional memory requirements. Therefore, the choice of expansion rate may be heavily dependent on the available hardware configuration. We leave a detailed study of these factors for future work.


\paragraph{Modeling the cost of granularity.}
It is important to note that the exact estimation of the training cost of MoE models is dependent on the training setup, hardware, and implementation. Specifically, increasing G can lead to higher transfer costs, depending on the adopted model of distributed training. Therefore, the precise selection of hyperparameters should be made considering these factors. In this work, we model the cost of operations using FLOPs, which is common in the Scaling Laws literature \citep{kaplan2020scaling, hoffmann2022training, frantar2023scaling}. Additionally, we would like to note that in our setup, we observe significant gains of fine-grained MoE measured as wall-clock time needed to achieve given perplexity (see Fig. \ref{fig:valid_wall} (b) for an example).  




\section{Conclusions}

This study introduces a novel hyperparameter, granularity ($G$), and underscores the significance of adjusting it for optimizing the efficiency of experts within MoE models.  A central finding of this research is that a standard granularity of $G=1$ is suboptimal across a broad range of FLOPs, leading to the recommendation of using higher granularity values to enhance MoE model performance and efficiency. Simultaneously, this work emphasizes the importance of varying training duration for compute-optimal settings. Consequently, both granularity and variable training length are incorporated into new scaling laws. These laws confidently demonstrate that MoE models consistently outperform dense transformers in terms of efficiency and scaling. This work not only sheds new light on the scaling laws applicable to MoE models but also provides practical guidance for improving computational efficiency in large language models. The insights are critical for the development and optimization of large-scale language models, marking a significant advancement in the field.

\section{Reproducibility}

The code used to produce the results described in this work is open-sourced and can be found at \href{https://github.com/llm-random/llm-random}{\texttt{github.com/llm-random/llm-random}}. 

\newpage
\section*{Acknowledgments}

We would like to express sincere gratitude to Piotr Miło\'s and Tomasz Trzci\'nski for valuable feedback and to Aleksandra Weglarz for her help with graphic design.

This work was funded by IDEAS NCBR, which also provided significant computational resources a supportive research environment and direction. The research was supported by PL-Grid infrastructure (grant PLG/2023/016148). We also benefited from the Entropy cluster (hosted at the Faculty of Mathematics, Informatics and Mechanics of the University of Warsaw) funded by NVIDIA, Intel, the Polish National Science Center grant 2022/45/N/ST6/02222, and ERC Starting Grant TOTAL. Marek Cygan was partially supported by an NCBiR grant POIR.01.01.01-00-0392/17-00.


\bibliography{iclr2024_conference}

\begin{thebibliography}{37}
\providecommand{\natexlab}[1]{#1}
\providecommand{\url}[1]{\texttt{#1}}
\expandafter\ifx\csname urlstyle\endcsname\relax
  \providecommand{\doi}[1]{doi: #1}\else
  \providecommand{\doi}{doi: \begingroup \urlstyle{rm}\Url}\fi

\bibitem[Agostinelli et~al.(2023)Agostinelli, Denk, Borsos, Engel, Verzetti, Caillon, Huang, Jansen, Roberts, Tagliasacchi, Sharifi, Zeghidour, and Frank]{agostinelli2023musiclm}
Andrea Agostinelli, Timo~I. Denk, Zalán Borsos, Jesse Engel, Mauro Verzetti, Antoine Caillon, Qingqing Huang, Aren Jansen, Adam Roberts, Marco Tagliasacchi, Matt Sharifi, Neil Zeghidour, and Christian Frank.
\newblock Musiclm: Generating music from text, 2023.

\bibitem[Artetxe et~al.(2022)Artetxe, Bhosale, Goyal, Mihaylov, Ott, Shleifer, Lin, Du, Iyer, Pasunuru, Anantharaman, Li, Chen, Akin, Baines, Martin, Zhou, Koura, O'Horo, Wang, Zettlemoyer, Diab, Kozareva, and Stoyanov]{artetxe2022efficient}
Mikel Artetxe, Shruti Bhosale, Naman Goyal, Todor Mihaylov, Myle Ott, Sam Shleifer, Xi~Victoria Lin, Jingfei Du, Srinivasan Iyer, Ramakanth Pasunuru, Giri Anantharaman, Xian Li, Shuohui Chen, Halil Akin, Mandeep Baines, Louis Martin, Xing Zhou, Punit~Singh Koura, Brian O'Horo, Jeff Wang, Luke Zettlemoyer, Mona Diab, Zornitsa Kozareva, and Ves Stoyanov.
\newblock Efficient large scale language modeling with mixtures of experts, 2022.

\bibitem[Brent(1971)]{Brent1971AnAW}
Richard~P. Brent.
\newblock An algorithm with guaranteed convergence for finding a zero of a function.
\newblock \emph{Comput. J.}, 14:\penalty0 422--425, 1971.
\newblock URL \url{https://api.semanticscholar.org/CorpusID:10312755}.

\bibitem[Brown et~al.(2020)Brown, Mann, Ryder, Subbiah, Kaplan, Dhariwal, Neelakantan, Shyam, Sastry, Askell, Agarwal, Herbert-Voss, Krueger, Henighan, Child, Ramesh, Ziegler, Wu, Winter, Hesse, Chen, Sigler, Litwin, Gray, Chess, Clark, Berner, McCandlish, Radford, Sutskever, and Amodei]{brown2020language}
Tom~B. Brown, Benjamin Mann, Nick Ryder, Melanie Subbiah, Jared Kaplan, Prafulla Dhariwal, Arvind Neelakantan, Pranav Shyam, Girish Sastry, Amanda Askell, Sandhini Agarwal, Ariel Herbert-Voss, Gretchen Krueger, Tom Henighan, Rewon Child, Aditya Ramesh, Daniel~M. Ziegler, Jeffrey Wu, Clemens Winter, Christopher Hesse, Mark Chen, Eric Sigler, Mateusz Litwin, Scott Gray, Benjamin Chess, Jack Clark, Christopher Berner, Sam McCandlish, Alec Radford, Ilya Sutskever, and Dario Amodei.
\newblock Language models are few-shot learners, 2020.

\bibitem[Chowdhery et~al.(2022)Chowdhery, Narang, Devlin, Bosma, Mishra, Roberts, Barham, Chung, Sutton, Gehrmann, Schuh, Shi, Tsvyashchenko, Maynez, Rao, Barnes, Tay, Shazeer, Prabhakaran, Reif, Du, Hutchinson, Pope, Bradbury, Austin, Isard, Gur-Ari, Yin, Duke, Levskaya, Ghemawat, Dev, Michalewski, Garcia, Misra, Robinson, Fedus, Zhou, Ippolito, Luan, Lim, Zoph, Spiridonov, Sepassi, Dohan, Agrawal, Omernick, Dai, Pillai, Pellat, Lewkowycz, Moreira, Child, Polozov, Lee, Zhou, Wang, Saeta, Diaz, Firat, Catasta, Wei, Meier-Hellstern, Eck, Dean, Petrov, and Fiedel]{chowdhery2022palm}
Aakanksha Chowdhery, Sharan Narang, Jacob Devlin, Maarten Bosma, Gaurav Mishra, Adam Roberts, Paul Barham, Hyung~Won Chung, Charles Sutton, Sebastian Gehrmann, Parker Schuh, Kensen Shi, Sasha Tsvyashchenko, Joshua Maynez, Abhishek Rao, Parker Barnes, Yi~Tay, Noam Shazeer, Vinodkumar Prabhakaran, Emily Reif, Nan Du, Ben Hutchinson, Reiner Pope, James Bradbury, Jacob Austin, Michael Isard, Guy Gur-Ari, Pengcheng Yin, Toju Duke, Anselm Levskaya, Sanjay Ghemawat, Sunipa Dev, Henryk Michalewski, Xavier Garcia, Vedant Misra, Kevin Robinson, Liam Fedus, Denny Zhou, Daphne Ippolito, David Luan, Hyeontaek Lim, Barret Zoph, Alexander Spiridonov, Ryan Sepassi, David Dohan, Shivani Agrawal, Mark Omernick, Andrew~M. Dai, Thanumalayan~Sankaranarayana Pillai, Marie Pellat, Aitor Lewkowycz, Erica Moreira, Rewon Child, Oleksandr Polozov, Katherine Lee, Zongwei Zhou, Xuezhi Wang, Brennan Saeta, Mark Diaz, Orhan Firat, Michele Catasta, Jason Wei, Kathy Meier-Hellstern, Douglas Eck, Jeff Dean, Slav Petrov, and Noah Fiedel.
\newblock Palm: Scaling language modeling with pathways, 2022.

\bibitem[Clark et~al.(2022)Clark, de~las Casas, Guy, Mensch, Paganini, Hoffmann, Damoc, Hechtman, Cai, Borgeaud, van~den Driessche, Rutherford, Hennigan, Johnson, Millican, Cassirer, Jones, Buchatskaya, Budden, Sifre, Osindero, Vinyals, Rae, Elsen, Kavukcuoglu, and Simonyan]{clark2022unified}
Aidan Clark, Diego de~las Casas, Aurelia Guy, Arthur Mensch, Michela Paganini, Jordan Hoffmann, Bogdan Damoc, Blake Hechtman, Trevor Cai, Sebastian Borgeaud, George van~den Driessche, Eliza Rutherford, Tom Hennigan, Matthew Johnson, Katie Millican, Albin Cassirer, Chris Jones, Elena Buchatskaya, David Budden, Laurent Sifre, Simon Osindero, Oriol Vinyals, Jack Rae, Erich Elsen, Koray Kavukcuoglu, and Karen Simonyan.
\newblock Unified scaling laws for routed language models, 2022.

\bibitem[Dai et~al.(2024)Dai, Deng, Zhao, Xu, Gao, Chen, Li, Zeng, Yu, Wu, Xie, Li, Huang, Luo, Ruan, Sui, and Liang]{dai2024deepseekmoe}
Damai Dai, Chengqi Deng, Chenggang Zhao, R.~X. Xu, Huazuo Gao, Deli Chen, Jiashi Li, Wangding Zeng, Xingkai Yu, Y.~Wu, Zhenda Xie, Y.~K. Li, Panpan Huang, Fuli Luo, Chong Ruan, Zhifang Sui, and Wenfeng Liang.
\newblock Deepseekmoe: Towards ultimate expert specialization in mixture-of-experts language models, 2024.

\bibitem[Du et~al.(2022)Du, Huang, Dai, Tong, Lepikhin, Xu, Krikun, Zhou, Yu, Firat, Zoph, Fedus, Bosma, Zhou, Wang, Wang, Webster, Pellat, Robinson, Meier-Hellstern, Duke, Dixon, Zhang, Le, Wu, Chen, and Cui]{du2022glam}
Nan Du, Yanping Huang, Andrew~M. Dai, Simon Tong, Dmitry Lepikhin, Yuanzhong Xu, Maxim Krikun, Yanqi Zhou, Adams~Wei Yu, Orhan Firat, Barret Zoph, Liam Fedus, Maarten Bosma, Zongwei Zhou, Tao Wang, Yu~Emma Wang, Kellie Webster, Marie Pellat, Kevin Robinson, Kathleen Meier-Hellstern, Toju Duke, Lucas Dixon, Kun Zhang, Quoc~V Le, Yonghui Wu, Zhifeng Chen, and Claire Cui.
\newblock Glam: Efficient scaling of language models with mixture-of-experts, 2022.

\bibitem[Faiz et~al.(2024)Faiz, Kaneda, Wang, Osi, Sharma, Chen, and Jiang]{faiz2024llmcarbon}
Ahmad Faiz, Sotaro Kaneda, Ruhan Wang, Rita Osi, Prateek Sharma, Fan Chen, and Lei Jiang.
\newblock Llmcarbon: Modeling the end-to-end carbon footprint of large language models, 2024.

\bibitem[Fedus et~al.(2022)Fedus, Zoph, and Shazeer]{fedus2022switch}
William Fedus, Barret Zoph, and Noam Shazeer.
\newblock Switch transformers: Scaling to trillion parameter models with simple and efficient sparsity, 2022.

\bibitem[Frantar et~al.(2023)Frantar, Riquelme, Houlsby, Alistarh, and Evci]{frantar2023scaling}
Elias Frantar, Carlos Riquelme, Neil Houlsby, Dan Alistarh, and Utku Evci.
\newblock Scaling laws for sparsely-connected foundation models, 2023.

\bibitem[Ghorbani et~al.(2021)Ghorbani, Firat, Freitag, Bapna, Krikun, Garcia, Chelba, and Cherry]{ghorbani2021scaling}
Behrooz Ghorbani, Orhan Firat, Markus Freitag, Ankur Bapna, Maxim Krikun, Xavier Garcia, Ciprian Chelba, and Colin Cherry.
\newblock Scaling laws for neural machine translation, 2021.

\bibitem[Henighan et~al.(2020)Henighan, Kaplan, Katz, Chen, Hesse, Jackson, Jun, Brown, Dhariwal, Gray, Hallacy, Mann, Radford, Ramesh, Ryder, Ziegler, Schulman, Amodei, and McCandlish]{henighan2020scaling}
Tom Henighan, Jared Kaplan, Mor Katz, Mark Chen, Christopher Hesse, Jacob Jackson, Heewoo Jun, Tom~B. Brown, Prafulla Dhariwal, Scott Gray, Chris Hallacy, Benjamin Mann, Alec Radford, Aditya Ramesh, Nick Ryder, Daniel~M. Ziegler, John Schulman, Dario Amodei, and Sam McCandlish.
\newblock Scaling laws for autoregressive generative modeling, 2020.

\bibitem[Hochreiter \& Schmidhuber(1997)Hochreiter and Schmidhuber]{hochreiter1997long}
Sepp Hochreiter and J{\"u}rgen Schmidhuber.
\newblock Long short-term memory.
\newblock \emph{Neural computation}, 9\penalty0 (8):\penalty0 1735--1780, 1997.

\bibitem[Hoffmann et~al.(2022)Hoffmann, Borgeaud, Mensch, Buchatskaya, Cai, Rutherford, de~Las~Casas, Hendricks, Welbl, Clark, Hennigan, Noland, Millican, van~den Driessche, Damoc, Guy, Osindero, Simonyan, Elsen, Rae, Vinyals, and Sifre]{hoffmann2022training}
Jordan Hoffmann, Sebastian Borgeaud, Arthur Mensch, Elena Buchatskaya, Trevor Cai, Eliza Rutherford, Diego de~Las~Casas, Lisa~Anne Hendricks, Johannes Welbl, Aidan Clark, Tom Hennigan, Eric Noland, Katie Millican, George van~den Driessche, Bogdan Damoc, Aurelia Guy, Simon Osindero, Karen Simonyan, Erich Elsen, Jack~W. Rae, Oriol Vinyals, and Laurent Sifre.
\newblock Training compute-optimal large language models, 2022.

\bibitem[Huber(1964)]{huber_loss}
Peter~J. Huber.
\newblock {Robust Estimation of a Location Parameter}.
\newblock \emph{The Annals of Mathematical Statistics}, 35\penalty0 (1):\penalty0 73 -- 101, 1964.
\newblock \doi{10.1214/aoms/1177703732}.
\newblock URL \url{https://doi.org/10.1214/aoms/1177703732}.

\bibitem[Jiang et~al.(2023)Jiang, Sablayrolles, Mensch, Bamford, Chaplot, de~las Casas, Bressand, Lengyel, Lample, Saulnier, Lavaud, Lachaux, Stock, Scao, Lavril, Wang, Lacroix, and Sayed]{jiang2023mistral}
Albert~Q. Jiang, Alexandre Sablayrolles, Arthur Mensch, Chris Bamford, Devendra~Singh Chaplot, Diego de~las Casas, Florian Bressand, Gianna Lengyel, Guillaume Lample, Lucile Saulnier, Lélio~Renard Lavaud, Marie-Anne Lachaux, Pierre Stock, Teven~Le Scao, Thibaut Lavril, Thomas Wang, Timothée Lacroix, and William~El Sayed.
\newblock Mistral 7b, 2023.

\bibitem[Jiang et~al.(2024)Jiang, Sablayrolles, Roux, Mensch, Savary, Bamford, Chaplot, de~las Casas, Hanna, Bressand, Lengyel, Bour, Lample, Lavaud, Saulnier, Lachaux, Stock, Subramanian, Yang, Antoniak, Scao, Gervet, Lavril, Wang, Lacroix, and Sayed]{jiang2024mixtral}
Albert~Q. Jiang, Alexandre Sablayrolles, Antoine Roux, Arthur Mensch, Blanche Savary, Chris Bamford, Devendra~Singh Chaplot, Diego de~las Casas, Emma~Bou Hanna, Florian Bressand, Gianna Lengyel, Guillaume Bour, Guillaume Lample, Lélio~Renard Lavaud, Lucile Saulnier, Marie-Anne Lachaux, Pierre Stock, Sandeep Subramanian, Sophia Yang, Szymon Antoniak, Teven~Le Scao, Théophile Gervet, Thibaut Lavril, Thomas Wang, Timothée Lacroix, and William~El Sayed.
\newblock Mixtral of experts, 2024.

\bibitem[Kaplan et~al.(2020)Kaplan, McCandlish, Henighan, Brown, Chess, Child, Gray, Radford, Wu, and Amodei]{kaplan2020scaling}
Jared Kaplan, Sam McCandlish, Tom Henighan, Tom~B. Brown, Benjamin Chess, Rewon Child, Scott Gray, Alec Radford, Jeffrey Wu, and Dario Amodei.
\newblock Scaling laws for neural language models, 2020.

\bibitem[Lepikhin et~al.(2020)Lepikhin, Lee, Xu, Chen, Firat, Huang, Krikun, Shazeer, and Chen]{lepikhin2020gshard}
Dmitry Lepikhin, HyoukJoong Lee, Yuanzhong Xu, Dehao Chen, Orhan Firat, Yanping Huang, Maxim Krikun, Noam Shazeer, and Zhifeng Chen.
\newblock Gshard: Scaling giant models with conditional computation and automatic sharding, 2020.

\bibitem[Lewis et~al.(2021)Lewis, Bhosale, Dettmers, Goyal, and Zettlemoyer]{lewis2021base}
Mike Lewis, Shruti Bhosale, Tim Dettmers, Naman Goyal, and Luke Zettlemoyer.
\newblock Base layers: Simplifying training of large, sparse models, 2021.

\bibitem[Liu et~al.(2023)Liu, Dettmers, Lin, Stoyanov, and Li]{liu2023unified}
Zeyu~Leo Liu, Tim Dettmers, Xi~Victoria Lin, Veselin Stoyanov, and Xian Li.
\newblock Towards a unified view of sparse feed-forward network in pretraining large language model, 2023.

\bibitem[Loshchilov \& Hutter(2019)Loshchilov and Hutter]{loshchilov2019decoupled}
Ilya Loshchilov and Frank Hutter.
\newblock Decoupled weight decay regularization, 2019.

\bibitem[Puigcerver et~al.(2023)Puigcerver, Riquelme, Mustafa, and Houlsby]{puigcerver2023sparse}
Joan Puigcerver, Carlos Riquelme, Basil Mustafa, and Neil Houlsby.
\newblock From sparse to soft mixtures of experts, 2023.

\bibitem[Radford et~al.(2018{\natexlab{a}})Radford, Narasimhan, Salimans, and Sutskever]{radford2018improving}
Alec Radford, Karthik Narasimhan, Tim Salimans, and Ilya Sutskever.
\newblock Improving language understanding by generative pre-training.
\newblock 2018{\natexlab{a}}.

\bibitem[Radford et~al.(2018{\natexlab{b}})Radford, Wu, Child, Luan, Amodei, and Sutskever]{noauthororeditor}
Alec Radford, Jeffrey Wu, Rewon Child, David Luan, Dario Amodei, and Ilya Sutskever.
\newblock Language models are unsupervised multitask learners.
\newblock 2018{\natexlab{b}}.
\newblock URL \url{https://d4mucfpksywv.cloudfront.net/better-language-models/language-models.pdf}.

\bibitem[Rae et~al.(2022)Rae, Borgeaud, Cai, Millican, Hoffmann, Song, Aslanides, Henderson, Ring, Young, Rutherford, Hennigan, Menick, Cassirer, Powell, van~den Driessche, Hendricks, Rauh, Huang, Glaese, Welbl, Dathathri, Huang, Uesato, Mellor, Higgins, Creswell, McAleese, Wu, Elsen, Jayakumar, Buchatskaya, Budden, Sutherland, Simonyan, Paganini, Sifre, Martens, Li, Kuncoro, Nematzadeh, Gribovskaya, Donato, Lazaridou, Mensch, Lespiau, Tsimpoukelli, Grigorev, Fritz, Sottiaux, Pajarskas, Pohlen, Gong, Toyama, de~Masson~d'Autume, Li, Terzi, Mikulik, Babuschkin, Clark, de~Las~Casas, Guy, Jones, Bradbury, Johnson, Hechtman, Weidinger, Gabriel, Isaac, Lockhart, Osindero, Rimell, Dyer, Vinyals, Ayoub, Stanway, Bennett, Hassabis, Kavukcuoglu, and Irving]{rae2022scaling}
Jack~W. Rae, Sebastian Borgeaud, Trevor Cai, Katie Millican, Jordan Hoffmann, Francis Song, John Aslanides, Sarah Henderson, Roman Ring, Susannah Young, Eliza Rutherford, Tom Hennigan, Jacob Menick, Albin Cassirer, Richard Powell, George van~den Driessche, Lisa~Anne Hendricks, Maribeth Rauh, Po-Sen Huang, Amelia Glaese, Johannes Welbl, Sumanth Dathathri, Saffron Huang, Jonathan Uesato, John Mellor, Irina Higgins, Antonia Creswell, Nat McAleese, Amy Wu, Erich Elsen, Siddhant Jayakumar, Elena Buchatskaya, David Budden, Esme Sutherland, Karen Simonyan, Michela Paganini, Laurent Sifre, Lena Martens, Xiang~Lorraine Li, Adhiguna Kuncoro, Aida Nematzadeh, Elena Gribovskaya, Domenic Donato, Angeliki Lazaridou, Arthur Mensch, Jean-Baptiste Lespiau, Maria Tsimpoukelli, Nikolai Grigorev, Doug Fritz, Thibault Sottiaux, Mantas Pajarskas, Toby Pohlen, Zhitao Gong, Daniel Toyama, Cyprien de~Masson~d'Autume, Yujia Li, Tayfun Terzi, Vladimir Mikulik, Igor Babuschkin, Aidan Clark, Diego de~Las~Casas, Aurelia Guy, Chris Jones,
  James Bradbury, Matthew Johnson, Blake Hechtman, Laura Weidinger, Iason Gabriel, William Isaac, Ed~Lockhart, Simon Osindero, Laura Rimell, Chris Dyer, Oriol Vinyals, Kareem Ayoub, Jeff Stanway, Lorrayne Bennett, Demis Hassabis, Koray Kavukcuoglu, and Geoffrey Irving.
\newblock Scaling language models: Methods, analysis \& insights from training gopher, 2022.

\bibitem[Raffel et~al.(2023)Raffel, Shazeer, Roberts, Lee, Narang, Matena, Zhou, Li, and Liu]{raffel2023exploring}
Colin Raffel, Noam Shazeer, Adam Roberts, Katherine Lee, Sharan Narang, Michael Matena, Yanqi Zhou, Wei Li, and Peter~J. Liu.
\newblock Exploring the limits of transfer learning with a unified text-to-text transformer, 2023.

\bibitem[Roller et~al.(2021)Roller, Sukhbaatar, Szlam, and Weston]{roller2021hash}
Stephen Roller, Sainbayar Sukhbaatar, Arthur Szlam, and Jason Weston.
\newblock Hash layers for large sparse models, 2021.

\bibitem[Scao et~al.(2023)Scao, Fan, Akiki, Pavlick, Ilić, Hesslow, Castagné, Luccioni, Yvon, Gallé, Tow, Rush, Biderman, Webson, Ammanamanchi, Wang, Sagot, Muennighoff, del Moral, Ruwase, Bawden, Bekman, McMillan-Major, Beltagy, Nguyen, Saulnier, Tan, Suarez, Sanh, Laurençon, Jernite, Launay, Mitchell, Raffel, Gokaslan, Simhi, Soroa, Aji, Alfassy, Rogers, Nitzav, Xu, Mou, Emezue, Klamm, Leong, van Strien, Adelani, Radev, Ponferrada, Levkovizh, Kim, Natan, Toni, Dupont, Kruszewski, Pistilli, Elsahar, Benyamina, Tran, Yu, Abdulmumin, Johnson, Gonzalez-Dios, de~la Rosa, Chim, Dodge, Zhu, Chang, Frohberg, Tobing, Bhattacharjee, Almubarak, Chen, Lo, Werra, Weber, Phan, allal, Tanguy, Dey, Muñoz, Masoud, Grandury, Šaško, Huang, Coavoux, Singh, Jiang, Vu, Jauhar, Ghaleb, Subramani, Kassner, Khamis, Nguyen, Espejel, de~Gibert, Villegas, Henderson, Colombo, Amuok, Lhoest, Harliman, Bommasani, López, Ribeiro, Osei, Pyysalo, Nagel, Bose, Muhammad, Sharma, Longpre, Nikpoor, Silberberg, Pai, Zink, Torrent,
  Schick, Thrush, Danchev, Nikoulina, Laippala, Lepercq, Prabhu, Alyafeai, Talat, Raja, Heinzerling, Si, Taşar, Salesky, Mielke, Lee, Sharma, Santilli, Chaffin, Stiegler, Datta, Szczechla, Chhablani, Wang, Pandey, Strobelt, Fries, Rozen, Gao, Sutawika, Bari, Al-shaibani, Manica, Nayak, Teehan, Albanie, Shen, Ben-David, Bach, Kim, Bers, Fevry, Neeraj, Thakker, Raunak, Tang, Yong, Sun, Brody, Uri, Tojarieh, Roberts, Chung, Tae, Phang, Press, Li, Narayanan, Bourfoune, Casper, Rasley, Ryabinin, Mishra, Zhang, Shoeybi, Peyrounette, Patry, Tazi, Sanseviero, von Platen, Cornette, Lavallée, Lacroix, Rajbhandari, Gandhi, Smith, Requena, Patil, Dettmers, Baruwa, Singh, Cheveleva, Ligozat, Subramonian, Névéol, Lovering, Garrette, Tunuguntla, Reiter, Taktasheva, Voloshina, Bogdanov, Winata, Schoelkopf, Kalo, Novikova, Forde, Clive, Kasai, Kawamura, Hazan, Carpuat, Clinciu, Kim, Cheng, Serikov, Antverg, van~der Wal, Zhang, Zhang, Gehrmann, Mirkin, Pais, Shavrina, Scialom, Yun, Limisiewicz, Rieser, Protasov, Mikhailov,
  Pruksachatkun, Belinkov, Bamberger, Kasner, Rueda, Pestana, Feizpour, Khan, Faranak, Santos, Hevia, Unldreaj, Aghagol, Abdollahi, Tammour, HajiHosseini, Behroozi, Ajibade, Saxena, Ferrandis, McDuff, Contractor, Lansky, David, Kiela, Nguyen, Tan, Baylor, Ozoani, Mirza, Ononiwu, Rezanejad, Jones, Bhattacharya, Solaiman, Sedenko, Nejadgholi, Passmore, Seltzer, Sanz, Dutra, Samagaio, Elbadri, Mieskes, Gerchick, Akinlolu, McKenna, Qiu, Ghauri, Burynok, Abrar, Rajani, Elkott, Fahmy, Samuel, An, Kromann, Hao, Alizadeh, Shubber, Wang, Roy, Viguier, Le, Oyebade, Le, Yang, Nguyen, Kashyap, Palasciano, Callahan, Shukla, Miranda-Escalada, Singh, Beilharz, Wang, Brito, Zhou, Jain, Xu, Fourrier, Periñán, Molano, Yu, Manjavacas, Barth, Fuhrimann, Altay, Bayrak, Burns, Vrabec, Bello, Dash, Kang, Giorgi, Golde, Posada, Sivaraman, Bulchandani, Liu, Shinzato, de~Bykhovetz, Takeuchi, Pàmies, Castillo, Nezhurina, Sänger, Samwald, Cullan, Weinberg, Wolf, Mihaljcic, Liu, Freidank, Kang, Seelam, Dahlberg, Broad, Muellner,
  Fung, Haller, Chandrasekhar, Eisenberg, Martin, Canalli, Su, Su, Cahyawijaya, Garda, Deshmukh, Mishra, Kiblawi, Ott, Sang-aroonsiri, Kumar, Schweter, Bharati, Laud, Gigant, Kainuma, Kusa, Labrak, Bajaj, Venkatraman, Xu, Xu, Xu, Tan, Xie, Ye, Bras, Belkada, and Wolf]{workshop2023bloom}
Teven~Le Scao, Angela Fan, Christopher Akiki, Ellie Pavlick, Suzana Ilić, Daniel Hesslow, Roman Castagné, Alexandra~Sasha Luccioni, François Yvon, Matthias Gallé, Jonathan Tow, Alexander~M. Rush, Stella Biderman, Albert Webson, Pawan~Sasanka Ammanamanchi, Thomas Wang, Benoît Sagot, Niklas Muennighoff, Albert~Villanova del Moral, Olatunji Ruwase, Rachel Bawden, Stas Bekman, Angelina McMillan-Major, Iz~Beltagy, Huu Nguyen, Lucile Saulnier, Samson Tan, Pedro~Ortiz Suarez, Victor Sanh, Hugo Laurençon, Yacine Jernite, Julien Launay, Margaret Mitchell, Colin Raffel, Aaron Gokaslan, Adi Simhi, Aitor Soroa, Alham~Fikri Aji, Amit Alfassy, Anna Rogers, Ariel~Kreisberg Nitzav, Canwen Xu, Chenghao Mou, Chris Emezue, Christopher Klamm, Colin Leong, Daniel van Strien, David~Ifeoluwa Adelani, Dragomir Radev, Eduardo~González Ponferrada, Efrat Levkovizh, Ethan Kim, Eyal~Bar Natan, Francesco~De Toni, Gérard Dupont, Germán Kruszewski, Giada Pistilli, Hady Elsahar, Hamza Benyamina, Hieu Tran, Ian Yu, Idris Abdulmumin,
  Isaac Johnson, Itziar Gonzalez-Dios, Javier de~la Rosa, Jenny Chim, Jesse Dodge, Jian Zhu, Jonathan Chang, Jörg Frohberg, Joseph Tobing, Joydeep Bhattacharjee, Khalid Almubarak, Kimbo Chen, Kyle Lo, Leandro~Von Werra, Leon Weber, Long Phan, Loubna~Ben allal, Ludovic Tanguy, Manan Dey, Manuel~Romero Muñoz, Maraim Masoud, María Grandury, Mario Šaško, Max Huang, Maximin Coavoux, Mayank Singh, Mike Tian-Jian Jiang, Minh~Chien Vu, Mohammad~A. Jauhar, Mustafa Ghaleb, Nishant Subramani, Nora Kassner, Nurulaqilla Khamis, Olivier Nguyen, Omar Espejel, Ona de~Gibert, Paulo Villegas, Peter Henderson, Pierre Colombo, Priscilla Amuok, Quentin Lhoest, Rheza Harliman, Rishi Bommasani, Roberto~Luis López, Rui Ribeiro, Salomey Osei, Sampo Pyysalo, Sebastian Nagel, Shamik Bose, Shamsuddeen~Hassan Muhammad, Shanya Sharma, Shayne Longpre, Somaieh Nikpoor, Stanislav Silberberg, Suhas Pai, Sydney Zink, Tiago~Timponi Torrent, Timo Schick, Tristan Thrush, Valentin Danchev, Vassilina Nikoulina, Veronika Laippala, Violette
  Lepercq, Vrinda Prabhu, Zaid Alyafeai, Zeerak Talat, Arun Raja, Benjamin Heinzerling, Chenglei Si, Davut~Emre Taşar, Elizabeth Salesky, Sabrina~J. Mielke, Wilson~Y. Lee, Abheesht Sharma, Andrea Santilli, Antoine Chaffin, Arnaud Stiegler, Debajyoti Datta, Eliza Szczechla, Gunjan Chhablani, Han Wang, Harshit Pandey, Hendrik Strobelt, Jason~Alan Fries, Jos Rozen, Leo Gao, Lintang Sutawika, M~Saiful Bari, Maged~S. Al-shaibani, Matteo Manica, Nihal Nayak, Ryan Teehan, Samuel Albanie, Sheng Shen, Srulik Ben-David, Stephen~H. Bach, Taewoon Kim, Tali Bers, Thibault Fevry, Trishala Neeraj, Urmish Thakker, Vikas Raunak, Xiangru Tang, Zheng-Xin Yong, Zhiqing Sun, Shaked Brody, Yallow Uri, Hadar Tojarieh, Adam Roberts, Hyung~Won Chung, Jaesung Tae, Jason Phang, Ofir Press, Conglong Li, Deepak Narayanan, Hatim Bourfoune, Jared Casper, Jeff Rasley, Max Ryabinin, Mayank Mishra, Minjia Zhang, Mohammad Shoeybi, Myriam Peyrounette, Nicolas Patry, Nouamane Tazi, Omar Sanseviero, Patrick von Platen, Pierre Cornette,
  Pierre~François Lavallée, Rémi Lacroix, Samyam Rajbhandari, Sanchit Gandhi, Shaden Smith, Stéphane Requena, Suraj Patil, Tim Dettmers, Ahmed Baruwa, Amanpreet Singh, Anastasia Cheveleva, Anne-Laure Ligozat, Arjun Subramonian, Aurélie Névéol, Charles Lovering, Dan Garrette, Deepak Tunuguntla, Ehud Reiter, Ekaterina Taktasheva, Ekaterina Voloshina, Eli Bogdanov, Genta~Indra Winata, Hailey Schoelkopf, Jan-Christoph Kalo, Jekaterina Novikova, Jessica~Zosa Forde, Jordan Clive, Jungo Kasai, Ken Kawamura, Liam Hazan, Marine Carpuat, Miruna Clinciu, Najoung Kim, Newton Cheng, Oleg Serikov, Omer Antverg, Oskar van~der Wal, Rui Zhang, Ruochen Zhang, Sebastian Gehrmann, Shachar Mirkin, Shani Pais, Tatiana Shavrina, Thomas Scialom, Tian Yun, Tomasz Limisiewicz, Verena Rieser, Vitaly Protasov, Vladislav Mikhailov, Yada Pruksachatkun, Yonatan Belinkov, Zachary Bamberger, Zdeněk Kasner, Alice Rueda, Amanda Pestana, Amir Feizpour, Ammar Khan, Amy Faranak, Ana Santos, Anthony Hevia, Antigona Unldreaj, Arash Aghagol,
  Arezoo Abdollahi, Aycha Tammour, Azadeh HajiHosseini, Bahareh Behroozi, Benjamin Ajibade, Bharat Saxena, Carlos~Muñoz Ferrandis, Daniel McDuff, Danish Contractor, David Lansky, Davis David, Douwe Kiela, Duong~A. Nguyen, Edward Tan, Emi Baylor, Ezinwanne Ozoani, Fatima Mirza, Frankline Ononiwu, Habib Rezanejad, Hessie Jones, Indrani Bhattacharya, Irene Solaiman, Irina Sedenko, Isar Nejadgholi, Jesse Passmore, Josh Seltzer, Julio~Bonis Sanz, Livia Dutra, Mairon Samagaio, Maraim Elbadri, Margot Mieskes, Marissa Gerchick, Martha Akinlolu, Michael McKenna, Mike Qiu, Muhammed Ghauri, Mykola Burynok, Nafis Abrar, Nazneen Rajani, Nour Elkott, Nour Fahmy, Olanrewaju Samuel, Ran An, Rasmus Kromann, Ryan Hao, Samira Alizadeh, Sarmad Shubber, Silas Wang, Sourav Roy, Sylvain Viguier, Thanh Le, Tobi Oyebade, Trieu Le, Yoyo Yang, Zach Nguyen, Abhinav~Ramesh Kashyap, Alfredo Palasciano, Alison Callahan, Anima Shukla, Antonio Miranda-Escalada, Ayush Singh, Benjamin Beilharz, Bo~Wang, Caio Brito, Chenxi Zhou, Chirag Jain,
  Chuxin Xu, Clémentine Fourrier, Daniel~León Periñán, Daniel Molano, Dian Yu, Enrique Manjavacas, Fabio Barth, Florian Fuhrimann, Gabriel Altay, Giyaseddin Bayrak, Gully Burns, Helena~U. Vrabec, Imane Bello, Ishani Dash, Jihyun Kang, John Giorgi, Jonas Golde, Jose~David Posada, Karthik~Rangasai Sivaraman, Lokesh Bulchandani, Lu~Liu, Luisa Shinzato, Madeleine~Hahn de~Bykhovetz, Maiko Takeuchi, Marc Pàmies, Maria~A Castillo, Marianna Nezhurina, Mario Sänger, Matthias Samwald, Michael Cullan, Michael Weinberg, Michiel~De Wolf, Mina Mihaljcic, Minna Liu, Moritz Freidank, Myungsun Kang, Natasha Seelam, Nathan Dahlberg, Nicholas~Michio Broad, Nikolaus Muellner, Pascale Fung, Patrick Haller, Ramya Chandrasekhar, Renata Eisenberg, Robert Martin, Rodrigo Canalli, Rosaline Su, Ruisi Su, Samuel Cahyawijaya, Samuele Garda, Shlok~S Deshmukh, Shubhanshu Mishra, Sid Kiblawi, Simon Ott, Sinee Sang-aroonsiri, Srishti Kumar, Stefan Schweter, Sushil Bharati, Tanmay Laud, Théo Gigant, Tomoya Kainuma, Wojciech Kusa, Yanis
  Labrak, Yash~Shailesh Bajaj, Yash Venkatraman, Yifan Xu, Yingxin Xu, Yu~Xu, Zhe Tan, Zhongli Xie, Zifan Ye, Mathilde Bras, Younes Belkada, and Thomas Wolf.
\newblock Bloom: A 176b-parameter open-access multilingual language model, 2023.

\bibitem[Shazeer et~al.(2017)Shazeer, Mirhoseini, Maziarz, Davis, Le, Hinton, and Dean]{shazeer2017outrageously}
Noam Shazeer, Azalia Mirhoseini, Krzysztof Maziarz, Andy Davis, Quoc Le, Geoffrey Hinton, and Jeff Dean.
\newblock Outrageously large neural networks: The sparsely-gated mixture-of-experts layer, 2017.

\bibitem[Shazeer et~al.(2018)Shazeer, Cheng, Parmar, Tran, Vaswani, Koanantakool, Hawkins, Lee, Hong, Young, Sepassi, and Hechtman]{shazeer2018meshtensorflow}
Noam Shazeer, Youlong Cheng, Niki Parmar, Dustin Tran, Ashish Vaswani, Penporn Koanantakool, Peter Hawkins, HyoukJoong Lee, Mingsheng Hong, Cliff Young, Ryan Sepassi, and Blake Hechtman.
\newblock Mesh-tensorflow: Deep learning for supercomputers, 2018.

\bibitem[Touvron et~al.(2023{\natexlab{a}})Touvron, Lavril, Izacard, Martinet, Lachaux, Lacroix, Rozière, Goyal, Hambro, Azhar, Rodriguez, Joulin, Grave, and Lample]{touvron2023llama}
Hugo Touvron, Thibaut Lavril, Gautier Izacard, Xavier Martinet, Marie-Anne Lachaux, Timothée Lacroix, Baptiste Rozière, Naman Goyal, Eric Hambro, Faisal Azhar, Aurelien Rodriguez, Armand Joulin, Edouard Grave, and Guillaume Lample.
\newblock Llama: Open and efficient foundation language models, 2023{\natexlab{a}}.

\bibitem[Touvron et~al.(2023{\natexlab{b}})Touvron, Martin, Stone, Albert, Almahairi, Babaei, Bashlykov, Batra, Bhargava, Bhosale, Bikel, Blecher, Ferrer, Chen, Cucurull, Esiobu, Fernandes, Fu, Fu, Fuller, Gao, Goswami, Goyal, Hartshorn, Hosseini, Hou, Inan, Kardas, Kerkez, Khabsa, Kloumann, Korenev, Koura, Lachaux, Lavril, Lee, Liskovich, Lu, Mao, Martinet, Mihaylov, Mishra, Molybog, Nie, Poulton, Reizenstein, Rungta, Saladi, Schelten, Silva, Smith, Subramanian, Tan, Tang, Taylor, Williams, Kuan, Xu, Yan, Zarov, Zhang, Fan, Kambadur, Narang, Rodriguez, Stojnic, Edunov, and Scialom]{touvron2023llama2}
Hugo Touvron, Louis Martin, Kevin Stone, Peter Albert, Amjad Almahairi, Yasmine Babaei, Nikolay Bashlykov, Soumya Batra, Prajjwal Bhargava, Shruti Bhosale, Dan Bikel, Lukas Blecher, Cristian~Canton Ferrer, Moya Chen, Guillem Cucurull, David Esiobu, Jude Fernandes, Jeremy Fu, Wenyin Fu, Brian Fuller, Cynthia Gao, Vedanuj Goswami, Naman Goyal, Anthony Hartshorn, Saghar Hosseini, Rui Hou, Hakan Inan, Marcin Kardas, Viktor Kerkez, Madian Khabsa, Isabel Kloumann, Artem Korenev, Punit~Singh Koura, Marie-Anne Lachaux, Thibaut Lavril, Jenya Lee, Diana Liskovich, Yinghai Lu, Yuning Mao, Xavier Martinet, Todor Mihaylov, Pushkar Mishra, Igor Molybog, Yixin Nie, Andrew Poulton, Jeremy Reizenstein, Rashi Rungta, Kalyan Saladi, Alan Schelten, Ruan Silva, Eric~Michael Smith, Ranjan Subramanian, Xiaoqing~Ellen Tan, Binh Tang, Ross Taylor, Adina Williams, Jian~Xiang Kuan, Puxin Xu, Zheng Yan, Iliyan Zarov, Yuchen Zhang, Angela Fan, Melanie Kambadur, Sharan Narang, Aurelien Rodriguez, Robert Stojnic, Sergey Edunov, and Thomas
  Scialom.
\newblock Llama 2: Open foundation and fine-tuned chat models, 2023{\natexlab{b}}.

\bibitem[Yin et~al.(2023)Yin, Fu, Zhao, Li, Sun, Xu, and Chen]{yin2023survey}
Shukang Yin, Chaoyou Fu, Sirui Zhao, Ke~Li, Xing Sun, Tong Xu, and Enhong Chen.
\newblock A survey on multimodal large language models, 2023.

\bibitem[Zhou et~al.(2022)Zhou, Lei, Liu, Du, Huang, Zhao, Dai, Chen, Le, and Laudon]{zhou2022mixtureofexperts}
Yanqi Zhou, Tao Lei, Hanxiao Liu, Nan Du, Yanping Huang, Vincent Zhao, Andrew Dai, Zhifeng Chen, Quoc Le, and James Laudon.
\newblock Mixture-of-experts with expert choice routing, 2022.

\bibitem[Zhou et~al.(2023)Zhou, Du, Huang, Peng, Lan, Huang, Shakeri, So, Dai, Lu, Chen, Le, Cui, Laundon, and Dean]{zhou2023brainformers}
Yanqi Zhou, Nan Du, Yanping Huang, Daiyi Peng, Chang Lan, Da~Huang, Siamak Shakeri, David So, Andrew Dai, Yifeng Lu, Zhifeng Chen, Quoc Le, Claire Cui, James Laundon, and Jeff Dean.
\newblock Brainformers: Trading simplicity for efficiency, 2023.

\end{thebibliography}
\bibliographystyle{iclr2024_conference}

\appendix

\clearpage
\appendix
\onecolumn
\section{Architecture and Training Setup}\label{app:setup}
All of the models considered in this work are decoder-only Transformers trained on the C4 dataset \citep{raffel2023exploring}. We use GPT2 tokenizer \citep{radford2018improving}. Each batch consists of 0.5M tokens packed into 2048 sequences. Our optimizer is AdamW \citep{loshchilov2019decoupled}, with a weight decay of $0.1.$ In each training run, we use the maximum learning rate of $2\mathrm{e}{-4},$ with linear warmup for $1\%$ steps and cosine decay to $2\mathrm{e}{-5}.$ To improve stability, we initialize weights using the truncated normal distribution with reduced scale, as advised in \cite{fedus2022switch}. The models are trained using mixed precision; we always keep the attention mechanism and router in high precision. We assume the \textit{infinite data} regime, as the number of training tokens for any of the runs is less than the number of tokens in the corpus. We follow \cite{hoffmann2022training} and perform our analysis on the smoothed training loss.

In MoE, we use the Expert Choice routing algorithm, as it guarantees a balanced expert load without tuning additional hyperparameters. To maintain compatibility with autoregressive language modeling, we apply the recipe described in \cite{zhou2022mixtureofexperts}: tokens are grouped by position across different sequences. The group size is always set to $256.$ We match the number of FLOPs for MoE and dense models with the same $\dmodel$ (meaning we activate an average of $8\dmodel^2$ parameters per token in each MoE layer). In the router, softmax is performed over the expert dimension, while we choose tokens over the token dimension, as this leads to the best performance (as opposed to performing softmax over the token dimension). We put an additional layer normalization before the output of MoE layer. This gives a small improvement for standard MoE, but is crucial for the performance of models with $G>1.$

Table \ref{tab:moe} and Table \ref{tab:dense} list the considered architecture and training variants for dense and MoE models, respectively. 

\begin{table}[h] 
    \centering
    \caption{Architecture and training variants (MoE models).}
    \vspace{0.4cm}
    \begin{tabular}{ccccccc}
    \toprule
        \#parameters (nonemb) & $\dmodel$ & $\nblocks$ & $\nheads$ & $D$ (in \#tokens) & $G$\\
        \midrule
        64x3M  & 256 & 4 & 4 & 16B, 33B, 66B & 1, 2, 4, 8, 16  \\
        64x7M & 384 & 4 & 6 & 16B, 33B, 66B & 1, 2, 4, 8, 16 \\
        64x13M & 512 & 4 & 8 & 16B, 33B, 66B & 1, 2, 4, 8, 16 \\
        64x13M & 512 & 4 & 8 & 130B & 1, 2, 4 \\
        64x25M & 512 & 8 & 8 & 16B, 33B, & 1, 2, 4, 8, 16 \\
        64x25M & 512 & 8 & 8 & 66B & 1, 2, 4, 8 \\
        64x49M & 640 & 10 & 10 & 16B, 33B & 1, 2, 4, 8, 16 \\
        64x49M & 640 & 10 & 10 & 66B & 1, 2, 4 \\
        64x85M & 768 & 12 & 12 & 33B & 1, 2, 4 \\ \bottomrule
    \end{tabular}
    \label{tab:moe}
\end{table}

\begin{table}[h] \label{table_dense}
    \centering
    \caption{Architecture and training variants (dense models).}
    \vspace{0.4cm}
    \begin{tabular}{ccccccc}
    \toprule
        \#parameters (nonemb) & $\dmodel$ & $\nblocks$ & $\nheads$ & $D$ (in \#tokens) \\
        \midrule
        3M  & 256 & 4 & 4 & 16B, 24B, 33B, 66B \\
        6M & 256 & 8 & 4 & 16B, 24B, 33B, 66B \\
        13M & 512 & 4 & 8 & 16B, 24B, 33B, 66B \\
        25M & 512 & 8 & 8 & 16B, 24B, 33B, 66B \\
        49M & 640 & 10 & 10 & 16B, 24B, 33B, 66B \\
        85M & 768 & 12 & 12 & 16B, 33B \\
        \bottomrule
    \end{tabular}
    \label{tab:dense}
\end{table}

\newpage
\section{Validation of the Scaling Law}\label{app:coeffs_valid}
In this section, we provide coefficients of the scaling law fitted with 20$\%$ of datapoints with the lowest perplexity excluded for the purpose of validation.

\begin{table}[h]
    \centering
    \caption{Values of the fitted coefficients.}
    \vspace{0.4cm}
    \begin{tabular}{c|c|c|c|c|c|c|c|c}
        Model  & a & $\alpha$ & b & $\beta$ & g & $\gamma$ & c \\
        \hline
        MoE  & 17.6 & 0.114 & 26.7 & 0.140 & 2.07 & 0.570 & 0.472 \\ 
    \end{tabular}
    \label{tab:coeffs_valid}
\end{table}

\section{Reliability of Compute Optimal Formula}\label{app:reliabiliy}
In this section, we assess the stability of our predictions presented in Section~\ref{sub:compute_optimal}. Similarly to \cite{hoffmann2022training} we calculate the 10$^\text{th}$ and 90$^\text{th}$ percentiles estimated via bootstrapping data ($80\%$ of the data is sampled $100$ times). See Table \ref{tab:bootstrap} for the details.

\begin{table}[h]
    \centering
    \caption{10$^\text{th}$ and 90$^\text{th}$ percentiles estimated via bootstraping data.}
    \vspace{0.4cm}
    \begin{tabular}{ccc}
        \toprule
        N & D & G \\
        \midrule
        64 x 100M & (2.97B, 5.98B) & (8, 8) \\ 
        64 x 1B & (21.17B, 40.73B)  & (16, 16)  \\
        64 x 3B & (50.20B, 105.88B)  &  (16, 32)  \\
        64 x 7B & (101.06B, 205.40B)   &  (32, 32)  \\
        64 x 70B & (638.49B, 1.59T) & (32, 64) \\
        64 x 300B & (1.99T, 5.62T)  &  (64, 64) \\
        64 x 1T & (5.29T, 16.87T)   & (64, 64) \\
        \bottomrule
    \end{tabular}
    \label{tab:bootstrap}
\end{table}
\section{Varying Expansion Rate} \label{app:exp_rate}
In this section, we provide results for $E=16.$ The training procedure is the same as described in App. \ref{app:setup}. The models considered in this part are listed in Table \ref{tab:moe_16}. 

\begin{table}[h] 
    \centering
    \caption{Architecture and training variants (MoE models).}
    \vspace{0.4cm}
    \begin{tabular}{ccccccc}
        \toprule
        \#parameters (nonemb) & $\dmodel$ & $\nblocks$ & $\nheads$ & $D$ (in \#tokens) & $G$\\
        \midrule
        64x3M  & 256 & 4 & 4 & 8B, 16B, 33B & 1, 2, 4, 8, 16  \\
        64x7M & 256 & 8 & 4 & 8B, 16B, 33B & 1, 2, 4, 8, 16 \\
        64x13M & 512 & 4 & 8 & 8B, 16B, 33B & 1, 2, 4, 8, 16 \\
        64x13M & 512 & 4 & 8 & 66B & 1, 2, 4 \\
        64x25M & 512 & 8 & 8 & 8B, 16B, 33B & 1, 2, 4, 8, 16 \\
        64x49M & 640 & 10 & 10 & 8B & 1, 2, 4, 8, 16 \\
        \bottomrule
    \end{tabular}
    \label{tab:moe_16}
\end{table}

We fit Eq.~\ref{eq:loss_moe} using the same procedure as described in Section~\ref{sub:parametric}. The results are detailed in Table \ref{tab:_different_e}.

\begin{table}[ht!]
    \centering
        \caption{Values of the fitted coefficients.}
    \vspace{0.4cm}
    \begin{tabular}{ccccccccc}
    
    \toprule
        Model  & a & $\alpha$ & b & $\beta$ & g & $\gamma$ & c \\
        \midrule
        MoE ($E=16$)  & 19.64 & 0.124 & 57.07 & 0.169 & 1.18 & 0.986 & 0.472 \\  \bottomrule
    \end{tabular}
    \label{tab:_different_e}
\end{table}

Using the coefficients and FLOPs calculation formulas, we can derive the compute optimal training parameters. The results are presented in Table \ref{tab:compute_opt_16}.

\begin{table}[ht!]
    \caption{10$^\text{th}$ and 90$^\text{th}$ percentiles estimated via bootstrapping data for $E=16$.}
    \centering
    \vspace{0.4cm}
    \begin{tabular}{ccc}
    \toprule
        N & D & G \\
        \midrule
        16 x 100M & (10.29B, 17.73B)   & (8 , 16)  \\
        16 x 1B   & (53.74B, 103.54B)  & (16, 32)  \\
        16 x 3B   & (106.22B, 261.04B) & (16, 32)  \\
        16 x 7B   & (177.65B, 511.43B) & (16, 32)  \\
        16 x 70B  & (721.60B, 3.22T)   & (32, 64)  \\
        16 x 300B & (1.73T, 10.69T)    & (32, 64)  \\
        16 x 1T   & (3.60T, 28.22T)    & (32, 128)  \\ \bottomrule
    \end{tabular}
    \label{tab:compute_opt_16}
\end{table}

We can observe that similarly to the case when $E=64,$ larger compute budgets imply larger optimal values of $G.$ Note that the values for $10^\text{th}$ and $90^\text{th}$ percentiles form larger intervals in this case, as in this part we run a smaller number of experiments and keep shorter training durations. However, we believe that this preliminary study forms a valuable addition to the results in the main part.

\section{FLOPs Constants}\label{app:constants}

The number of FLOPs $F$ used in Transformer training, considering the routing operation overhead in MoE, can be described by the following formula:
\begin{align}
     F = (12{\dmodel}^2c_f + \dmodel EGc_r) \cdot \ntokens \cdot \nlayers
    \label{eq:flops_overhead}
\end{align}

Following \cite{hoffmann2022training}, we assume $c_f$ to be $6$. This is interpreted as 6 FLOPs for each pair of an active parameter (in linear projection) and a processed token. The breakdown of operations is as follows:
\begin{itemize}
    \item During the forward pass, 2 operations (single multiplication and single addition) are used to compute the matrix multiplication of an input and linear projection.
    \item During the backward pass, 2 operations are used to compute gradients wrt. the input.
    \item During the backward pass, 2 operations are used to compute gradients wrt. the weights of linear projection.
\end{itemize}

In our work, we have assumed the routing constant, $c_r$, to be 14, with the breakdown presented below. The exact number of operations may depend on the implementation of routing, but it will be between 6 and 20. However, our main conclusions of the paper are resistant to different assumptions of this constant.
\begin{itemize}
    \item During the forward pass, 2 operations are used to compute the expert logits based on an input and ``routing linear projection''.
    \item During the backward pass, 2 operations are used to compute gradients for ``routing linear projection'' wrt. the input.
    \item During the backward pass, 2 operations are used to compute gradients for ``routing linear projection'' wrt. the weights of linear projection.
    \item During the forward pass, 2 operations are used to route input tokens to chosen experts.
    \item During the forward pass, 2 operations are used to route expert outputs to chosen tokens and multiply those outputs by the routing score.
    \item During the backward pass, 2 operations are used to route gradients from output tokens to experts.
    \item During the backward pass, 2 operations are used to route gradients from experts to input tokens.
\end{itemize}

Similarly to the calculation of FLOPs for $c_f$, FLOPs come in pairs as each multiplication is followed by an addition (used to accumulate outputs or gradients).

\section{Additional Visualizations}
\label{sec:app_add_viz}

\begin{figure}[ht!]
  \centering
  \begin{tabular}{ c @{\hspace{30pt}} c }
    \includegraphics[width=.3\textwidth]{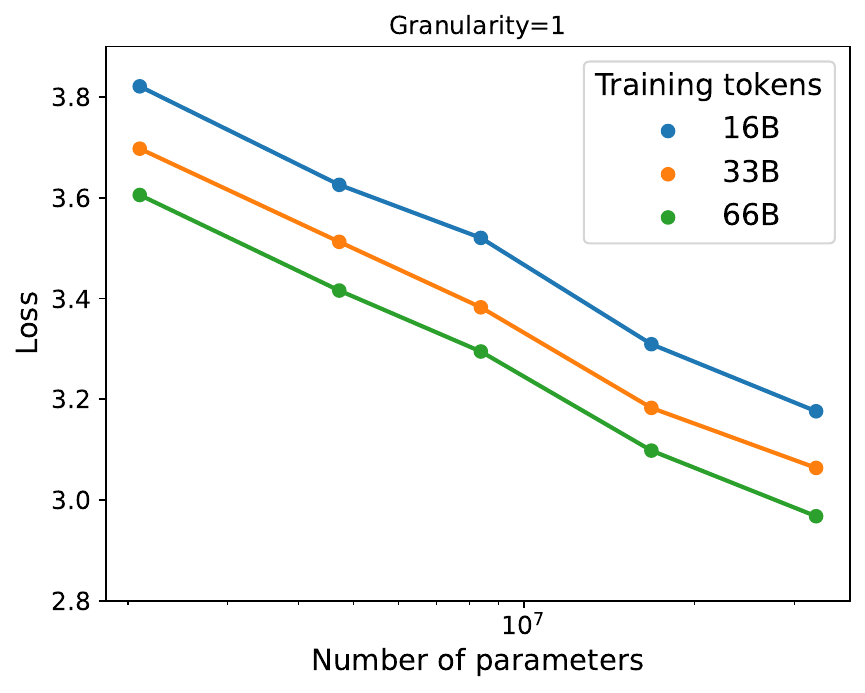} &
      \includegraphics[width=.3\textwidth]{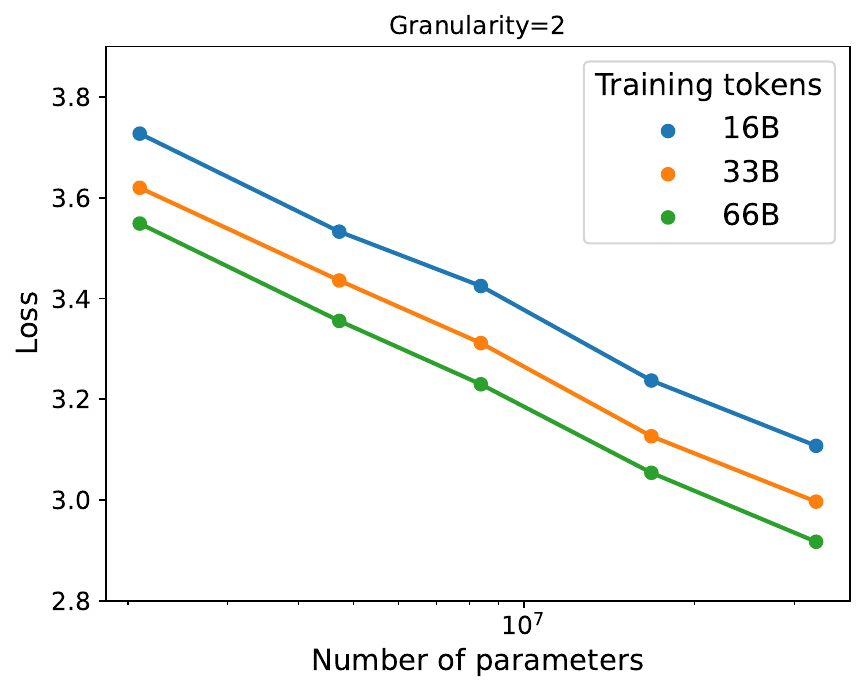} \\
    \small \textbf{(a)} &
      \small \textbf{(b)} \\
  \includegraphics[width=.3\textwidth]{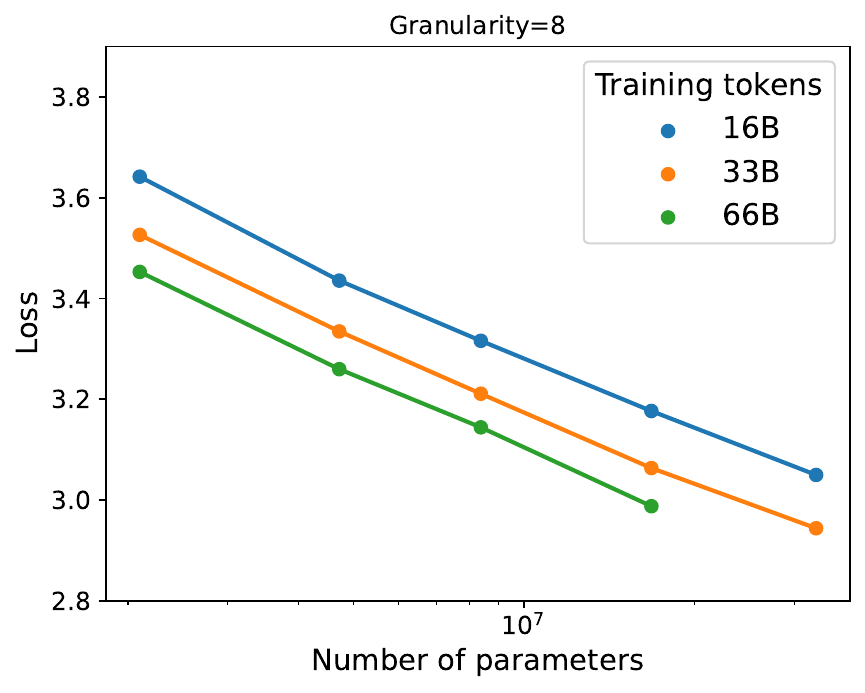} &
      \includegraphics[width=.3\textwidth]{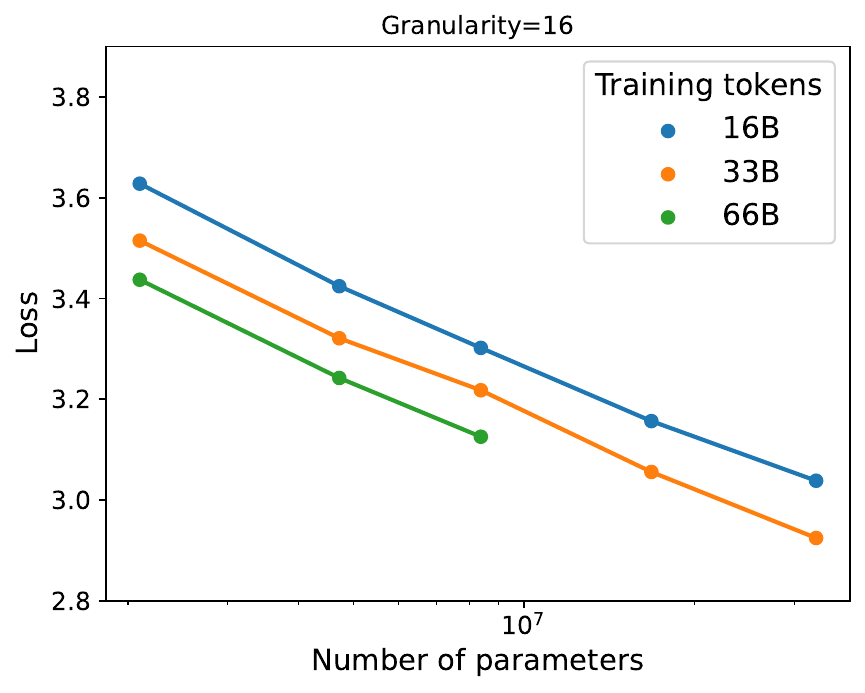} \\
    \small \textbf{(c)} &
      \small \textbf{(d)}
  \end{tabular}

  \medskip

  \caption{Illustration of scaling $N$ and $D$ for constant granularity value of: \textbf{(a)} $G=1$ \textbf{(b)} $G=2$ \textbf{(c)} $G=8$ \textbf{(d)} $G=16.$ }
  \label{line_gran_plot}
\end{figure}

\begin{figure}[ht!]
  \centering
  \begin{tabular}{ c @{\hspace{30pt}} c }
    \includegraphics[width=.3\textwidth]{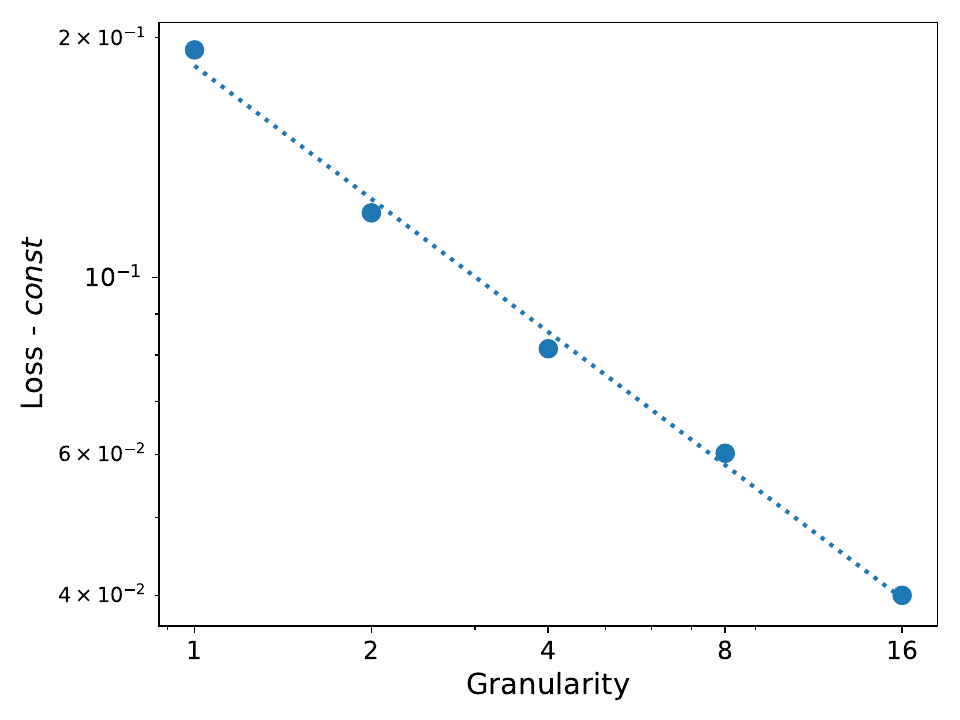} &
      \includegraphics[width=.3\textwidth]{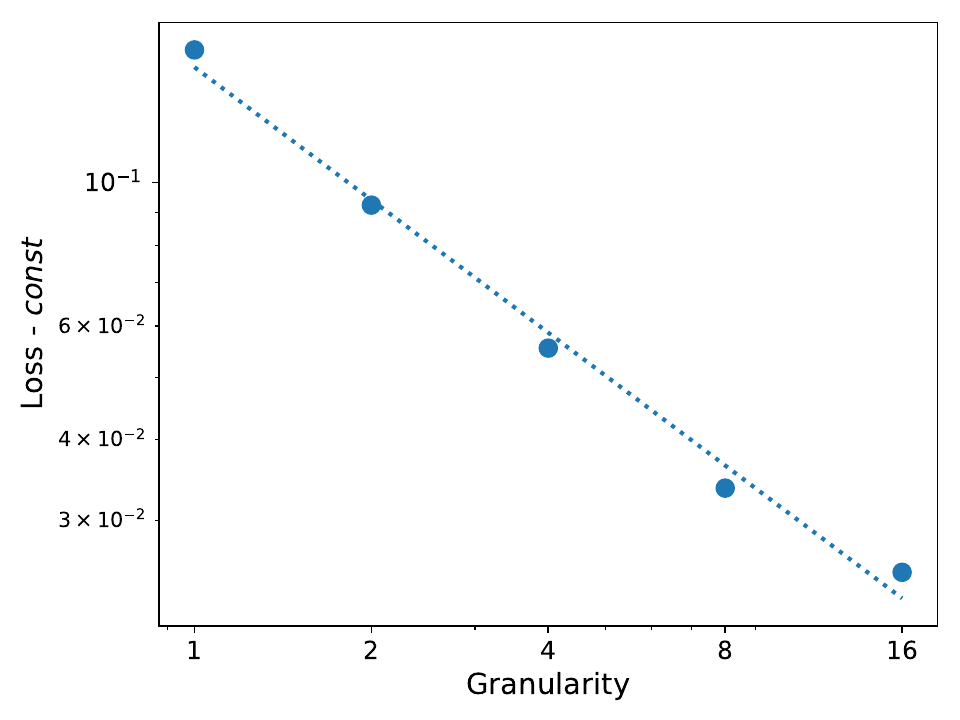} \\
    \small \textbf{(a)} &
      \small \textbf{(b)} \\
  \includegraphics[width=.3\textwidth]{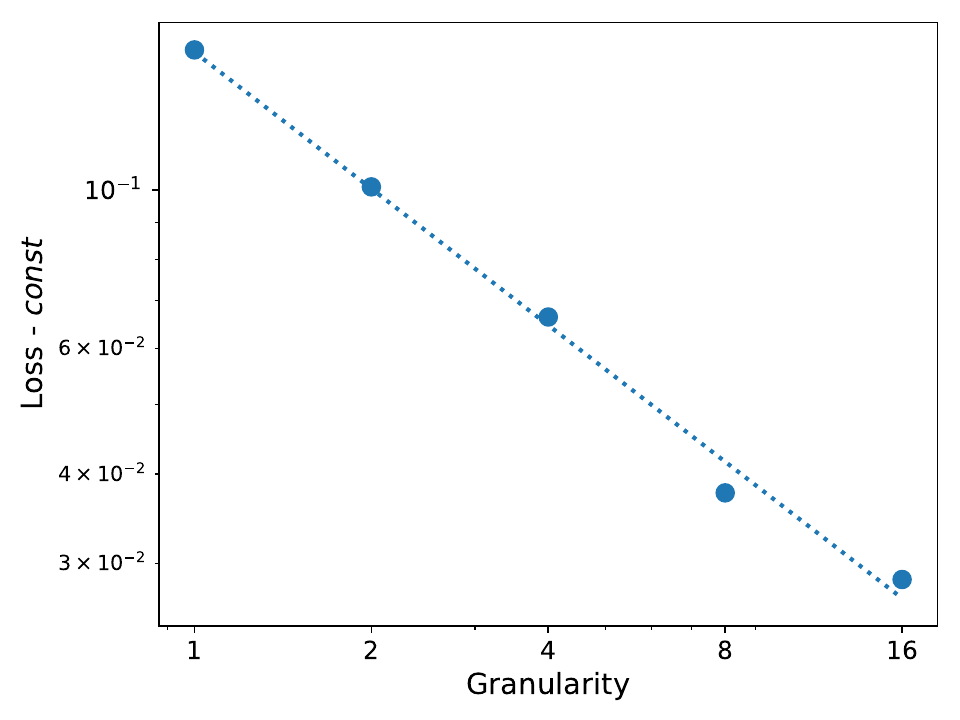} &
      \includegraphics[width=.3\textwidth]{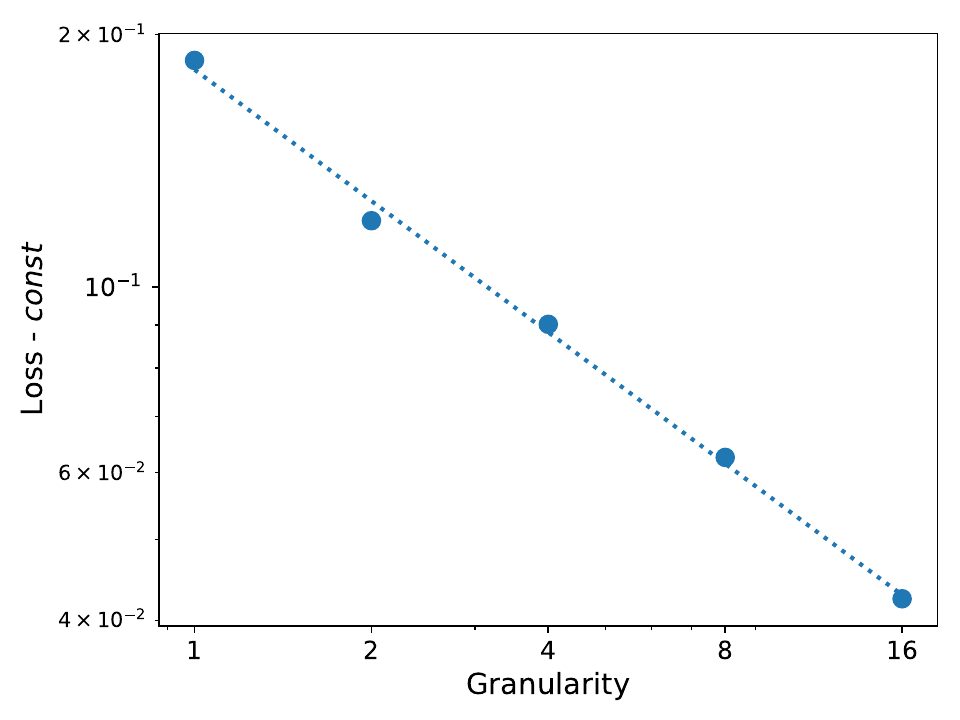} \\
    \small \textbf{(c)} &
      \small \textbf{(d)}
  \end{tabular}

  \medskip

  \caption{Illustration of scaling granularity when $N, D$ are fixed for: \textbf{(a)} $N=64 \times 25M$, $D=16B$, $const=3.12$ \textbf{(b)} $N=64 \times 49M $, $D=16B$, $const=3.02$ \textbf{(c)} $N=64 \times 25M$,  $D=32B$, $const=3.03$ \textbf{(d)} $N=64 \times 49M$, $D=32B$, $const=2.88$ }
  \label{line_gran_plot_fixed}
\end{figure}
\end{document}